\documentclass{article}

\usepackage{microtype}
\usepackage{graphicx}
\usepackage{subfigure}
\usepackage{booktabs} 
\usepackage{multirow}
\usepackage{xfrac}
\usepackage{diagbox}

\usepackage{float}

\usepackage{hyperref}

\usepackage[accepted]{icml2020}

\usepackage{amsmath}
\usepackage{amsfonts}
\usepackage{lipsum}
\usepackage{bbm}
\usepackage[utf8]{inputenc}
\usepackage[english]{babel}
\usepackage{amsthm}
\usepackage{xspace}

\newcommand{\bx}{\mathbf{x}}

\newcommand{\E}{\mathop{\mathbb{E}}}
\newcommand{\cL}{\mathcal{L}}
\newcommand{\cD}{\mathcal{D}}
\newcommand{\robin}{\textsc{RoBin}\xspace}

\usepackage{xargs}

\graphicspath{{./tmp_figures/}}

\newenvironment{itemize*}%
  {\vspace{-2ex} \begin{itemize} %
     \setlength{\itemsep}{-1ex} \setlength{\parsep}{0pt}}%
  {\end{itemize}}
  
\newenvironment{enumerate*}%
  {\vspace{-2ex} \begin{enumerate} %
     \setlength{\itemsep}{-1ex} \setlength{\parsep}{0pt}}%
  {\end{enumerate}}

\icmltitlerunning{Robustness from Simple Classifiers}

\begin{document}

\twocolumn[
\icmltitle{Robustness from Simple Classifiers}



\icmlsetsymbol{equal}{*}

\begin{icmlauthorlist}
\icmlauthor{Sharon Qian}{to}
\icmlauthor{Dimitris Kalimeris}{to}
\icmlauthor{Gal Kaplun}{to}
\icmlauthor{Yaron Singer}{to}
\end{icmlauthorlist}

\icmlaffiliation{to}{Harvard University, Cambridge, MA, USA}

\icmlcorrespondingauthor{}{sharonqian@g.harvard.edu}

\icmlkeywords{Machine Learning, ICML}

\vskip 0.3in
]



\printAffiliationsAndNotice{} 

\begin{abstract}

Despite the vast success of Deep Neural Networks in numerous application domains, it has been shown that such models are not robust i.e., they are vulnerable to small adversarial perturbations of the input. While extensive work has been done on why such perturbations occur or how to successfully defend against them, we still do not have a complete understanding of robustness. In this work we investigate the connection between robustness and simplicity. We find that simpler classifiers, formed by reducing the number of output classes, are less susceptible to adversarial perturbations. Consequently, we demonstrate that decomposing a complex multiclass model into an aggregation of binary models enhances robustness. This behavior is consistent across different datasets and model architectures and can be combined with known defense techniques such as adversarial training. Moreover, we provide further evidence of a disconnect between standard and robust learning regimes. In particular, we show that elaborate label information can help standard accuracy but harm robustness.

\end{abstract}

\section{Introduction}

Deep Neural Networks (DNNs) are vulnerable to adversarial examples, human-imperceptible perturbations of the inputs that are specifically crafted to fool the classifier \cite{szegedy2013intriguing, goodfellow}. This prevalent vulnerability is especially concerning as the usage of neural networks in industrial applications increases and the need for secure deep learning systems rapidly grows, e.g., in autonomous driving, online content moderation and malware detection \cite{synthesizing}.
 
In response to this vulnerability, a large body of work
has been devoted to constructing robust deep learning systems. While some progress has been made in understanding and defending against adversarial examples, most attempts have fallen to powerful optimization-based attacks \cite{masking, obfuscated}. The most successful and sustainable defense paradigm thus far is \emph{adversarial training} \cite{madry2017towards, goodfellow2014explaining} and its variants \cite{mma, trades, mart}. Since robustness could potentially be achieved using various methods, a natural question is whether we can design techniques complementary to adversarial training to decrease the vulnerability of our models.

\begin{figure}[t]
\begin{center}
\centerline{\includegraphics[width=0.85\columnwidth]{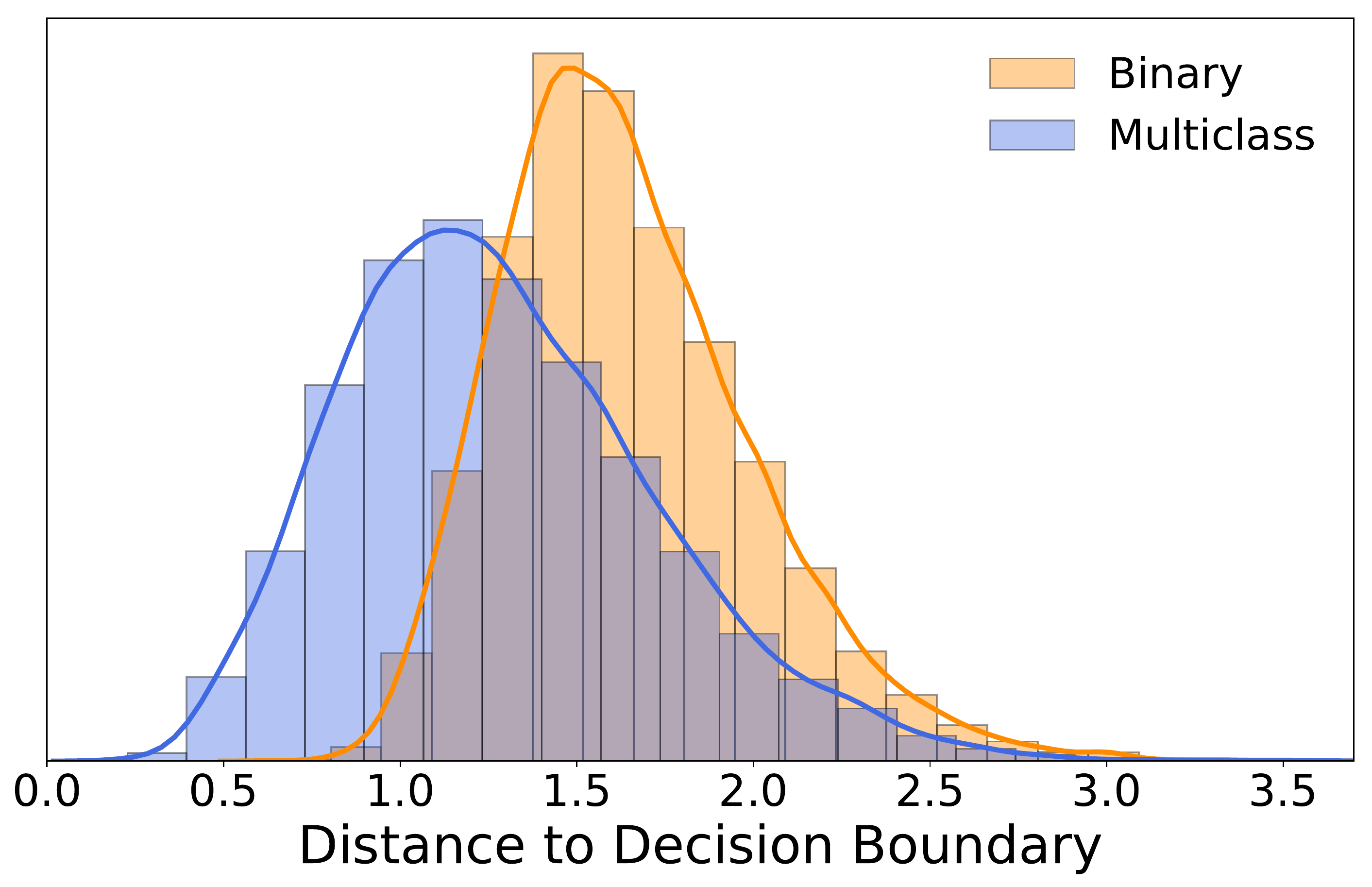}}
\caption{Distribution of distance to decision boundary of test examples for a ``simple" binary classifier (orange) and ``complex" multiclass classifier (blue). Smaller perturbations are sufficient to fool the multiclass classifier. Both models are ResNet-18s trained on CIFAR-10. (See Section \ref{sec:simprob} for details.)}
\label{fig:intro}
\end{center}
\vspace{-1cm}
\end{figure}

In this paper, we investigate this question by examining robustness through the lens of simplicity. 
More specifically, we consider the simplification of a classification task, or model, through the reduction of the number of its output classes. We then explore how the robust accuracy of the model evolves as a function of task simplicity.
The answer is unclear as there are two opposing intuitions: On one hand, tasks of lower complexity are easier to learn and therefore, simpler classifiers might be more robust. On the other hand, more classes or finer data labels provide more information to the model, which might lead to better performance.

In this context, we find that models trained on simpler objectives are less susceptible to adversarial perturbations. To measure robustness, we consider the distance between samples and decision boundaries for both binary and multiclass models. Figure~\ref{fig:intro} compares the distributions of these distances and shows a significant reduction in perturbation size when given a more complex classification task. Intuitively, this suggests that binary classifiers are more resilient to adversarial perturbations than their multiclass counterparts. 
%

The main obstacle is then how to leverage the observation above to develop multiclass models that enjoy an improvement in robustness.

\paragraph{Aggregation of Binary Classifiers.}  We propose an aggregation of adversarially trained binary classifiers. We measure the robustness of this model by crafting attacks specifically for our architecture. Surprisingly, this straightforward aggregation yields a consistent increase in robustness for the multiclass task when compared to the baseline of a single multiclass model. Finally, to understand this improvement, we analyze the correlation between the gradients of binary classifiers in the aggregation. We find that gradients are misaligned, making it difficult for an adversary to find effective perturbations using gradient-based attacks.

\paragraph{Contributions.} We make the following contributions:
\begin{itemize*}
\setlength\itemsep{0.005em}
    \item We propose a new connection between robustness and simplicity by showing that classifiers with fewer output classes are less susceptible to adversarial perturbations.
    \item We show that aggregations of simple classifiers have higher robust accuracy compared to multiclass models.
    \item We use gradient alignment analysis to understand the robustness of the aggregation and find that adversarial perturbations are not shared among its components. 
    \item Of independent interest, we demonstrate a surprising disconnect between learning a classifier for standard classification and a classifier robust to adversarial perturbations. Specifically, additional information on the dataset helps clean accuracy while harming robustness.
\end{itemize*}

\paragraph{Related Work.}

There have been numerous approaches proposed to defend against adversarial perturbations that yield a deeper understanding of robustness, including defensive distillation \cite{def_distil}, gradient regularization \cite{RDV18}, feature squeezing \cite{feat_sq} and model compression \cite{compression}. Moreover, \citet{detect1, detect2, detect18} focus on detecting adversarial noise.
Another line of defense strategies focuses on constructing ensembles of multiclass models to improve robustness \cite{tramer_ens}. 
While the early work of \citet{attackensembles} provides evidence that ensembles are, in general, not more robust than their weaker components, recent work show promising results by explicitly promoting diversity among individual classifiers \cite{diversity-training, promo-diversity}. 

\paragraph{Paper Organization.} In Section~\ref{sec:obj} we introduce preliminary definitions and notation. In Section~\ref{sec:simprob} we state and experimentally support our main observation about the robustness of simpler classifiers, which serves as motivation to introduce aggregations of binary classifiers in Section \ref{sec:algo}. We show empirical results regarding the performance of binary aggregations across different settings in Section~\ref{sec:exp} and provide intuition about the observed increase in accuracy using gradient alignment analysis in Section~\ref{sec:gradient}. We conclude with a discussion in Sections~\ref{sec:discussion} and \ref{sec:conclusion}. 

\begin{figure*}[t]
\begin{center}
\begin{tabular} {ccccc}
  \includegraphics[width=1.5in]{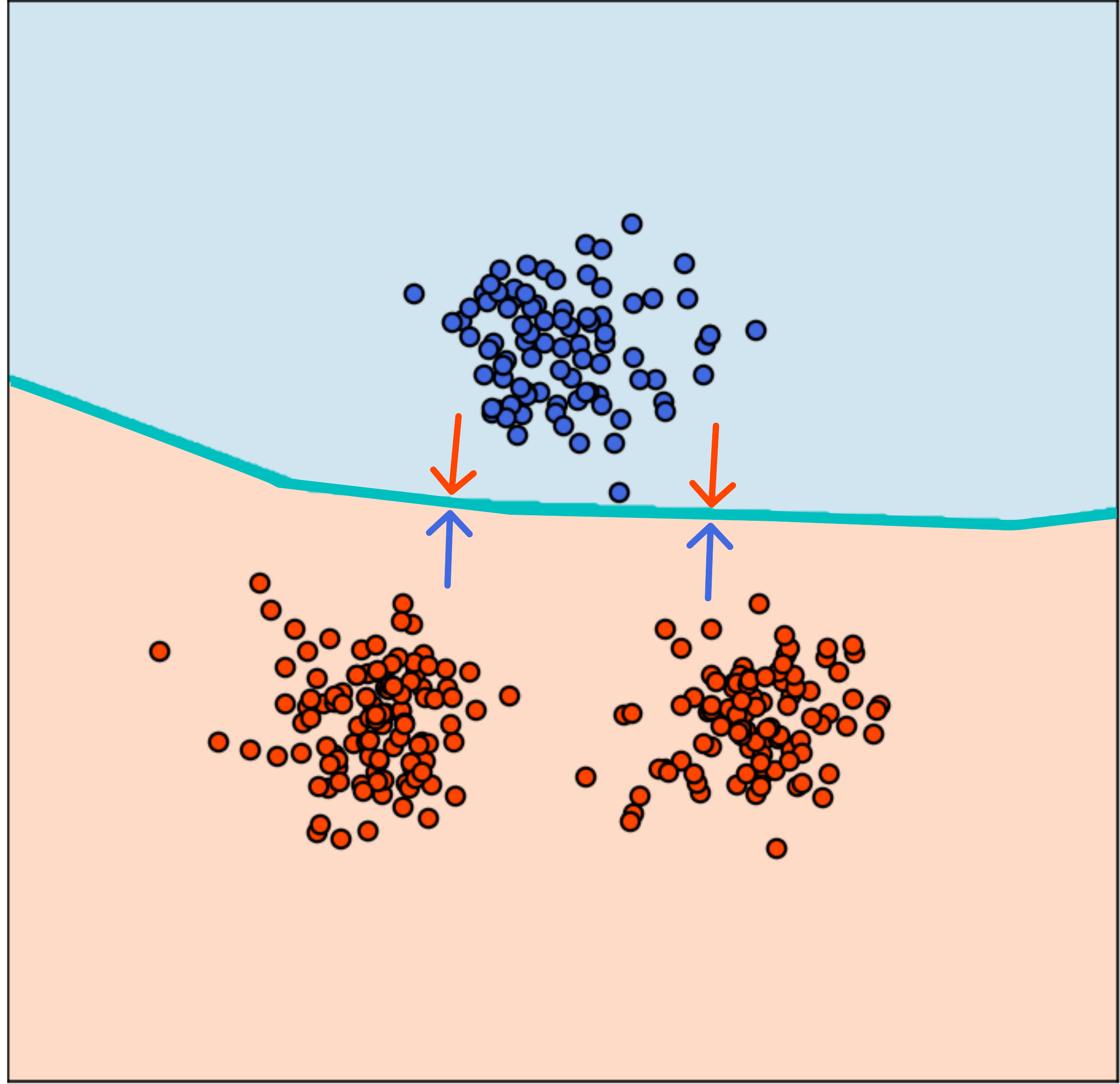} & \hspace{.05in} &
  \includegraphics[width=1.5in]{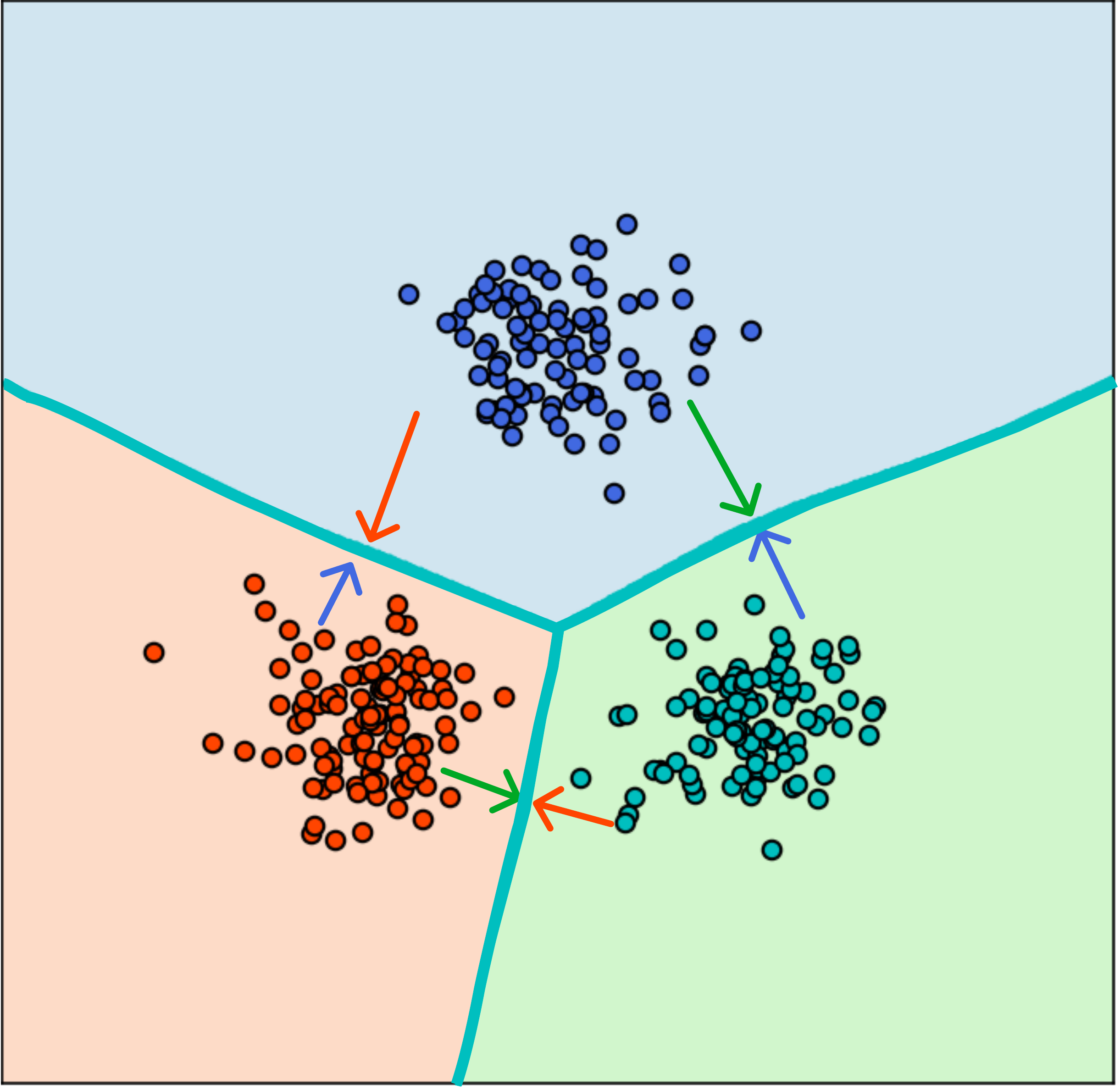} & \hspace{.05in} &
  \includegraphics[width=1.5in]{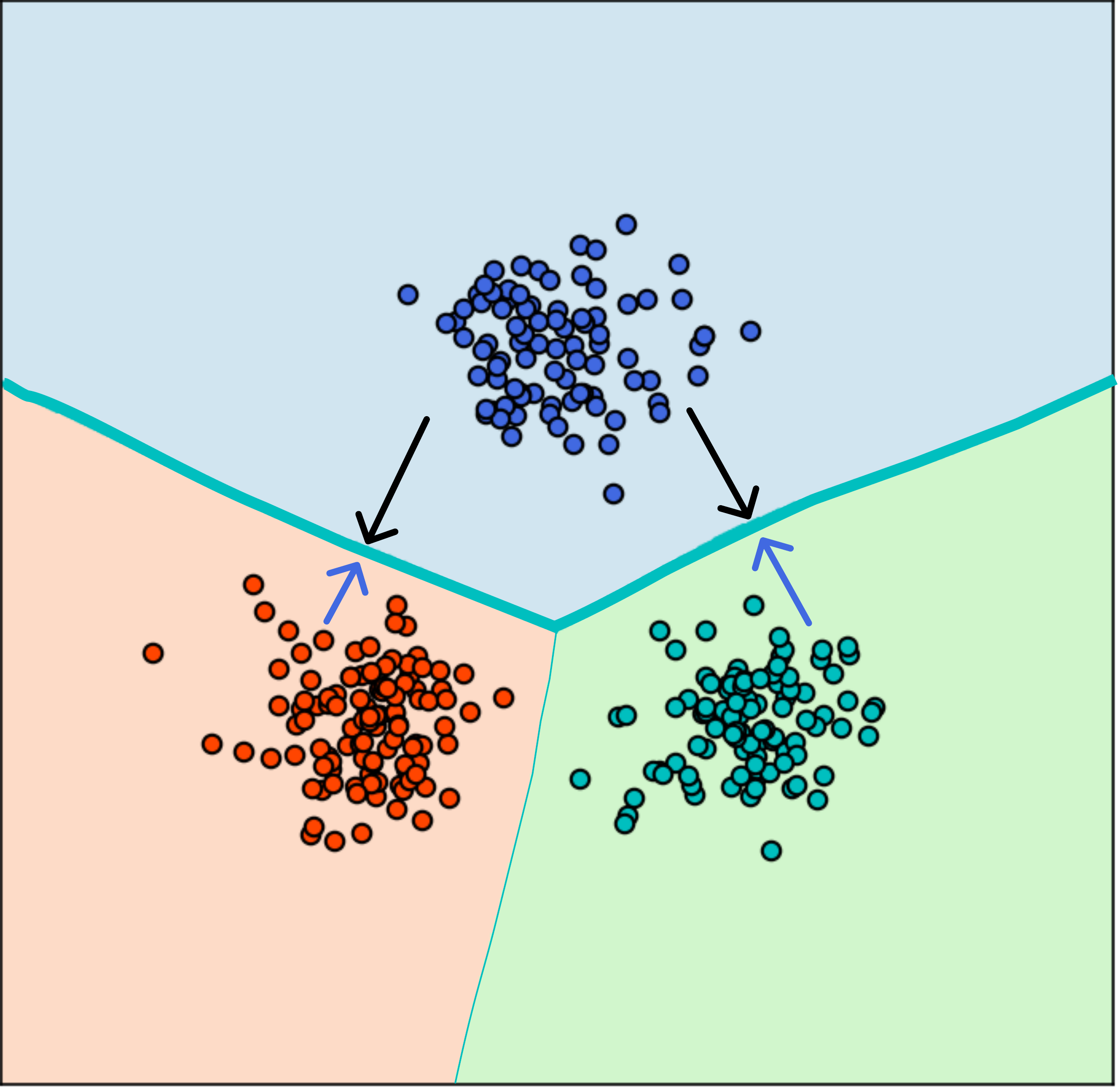} \\
  \end{tabular}
\end{center}
  \caption{Visualizations of learned decision boundaries for (a) a binary classifier and (b) a multiclass classifier that are given the same test time task but different train data labels. Robustness is measured using an untargeted adversary (see arrows) in (a) and (b). Robustness of the multiclass model in (c) is measured only in the context of the teal decision boundary to compare performance to the binary case.}

  \label{fig:no_noise}
\end{figure*}

\section{Preliminaries} \label{sec:obj}
In this section, we define our notation and briefly describe the adversarial attacks we use in our experiments.
\subsection{The Adversarial Setting}
\paragraph{Robust Optimization.} 
In the robust optimization framework of supervised learning the input examples $\bx\in \mathcal X$ and corresponding labels  $y \in \mathcal Y$  are generated from an unknown distribution $(\bx, y)\sim\cD$. For each natural input $\bx$, an adversary attempts to find an \emph{ adversarial example} by crafting noise $\xi$, so that the perturbed input $\bx' = \bx + \xi$ is close to $\bx$ but is misclassified by the model.

The goal of the learner is to output a classifier  $h_\theta: \mathcal X \to \mathcal{Y}$ 
that achieves low error on the adversarial distribution, i.e. is \emph{robust}. 
Here, $\theta$ denotes the weights of the model and we will omit it when it is clear from context. 
Formally,  we try to minimize the loss over adversarial examples:
    $$\min_\theta \E_{(\bx, y)\sim \cD} \max_{\bx'} \cL(h_\theta(\bx'), y)$$
where $\cL(h(\bx), y)$ is the cross-entropy loss. This optimization paradigm is known as \emph{adversarial training} and can be viewed as data augmentation with adversarial examples.

\paragraph{Threat Model.} 
In this paper we consider \emph{white-box} attacks where the adversary has full information about the architecture, weights and train data of the model. The only constraint is that the noise budget has bounded $\ell_p$ norm, ensuring that the adversarial example $\bx'$ lies in close proximity to the natural one. We consider both $\ell_2$ and $\ell_\infty$ attacks.

An attack is \emph{targeted} if the adversary crafts the perturbation so that $\bx'$ will be classified as an instance of a specific class while it is \emph{untargeted} if it aims to change the prediction of a model to any class other than the true class. We mostly consider untargeted attacks which is a more general setting.

\subsection{Attack Algorithms} \label{subsec:attack-alg}

\paragraph{Projected Gradient Descent (PGD).}
PGD was proposed by \citet{madry2017towards} as an improvement upon Iterative-FGSM \cite{bim}. It consists of two alternating steps: a gradient step in the direction maximizing the loss and a projection step constraining the distorted input to remain within the budget $\epsilon$. Formally,

\begin{align*}
  \bx^{k+1} = \Pi_\mathcal{B}(\bx^k + \eta \nabla_\bx \cL(y, h(\bx^k))  
\end{align*}

where $\mathcal B = \{ \bx^{k} \,:\, \| \bx^{k}-\bx\|_p \leq \epsilon\}$. We initiate $\bx_0$ to be a random point in the vicinity of $\bx$. As suggested in the original paper, we normalize the gradient for $p=2$ and use the sign of the gradient when $p=\infty$.
\paragraph{Carlini Wagner Attack (CW).} As in \citet{cw}, we use a similar approach to PGD and define $\bx(\omega) = \frac{1}{2}(\tanh(\omega) + 1)$ to avoid the projection step. Then, we minimize $\|\bx(\omega) - \bx\|+cf(\bx(\omega))$ to produce adversarial examples. Here, $f$ is a modified hinge loss and $c$ is tuned using binary search. While the core idea is the same for both $\ell_2$ and $\ell_\infty$ attacks, the implementation differs to ensure tractable optimization. 
\section{Robustness from Simplicity} \label{sec:simprob}

\begin{figure*}[ht]
\begin{center}
\begin{minipage}[t]{0.48\linewidth} 
    \centering
    \includegraphics[width=0.93\textwidth]{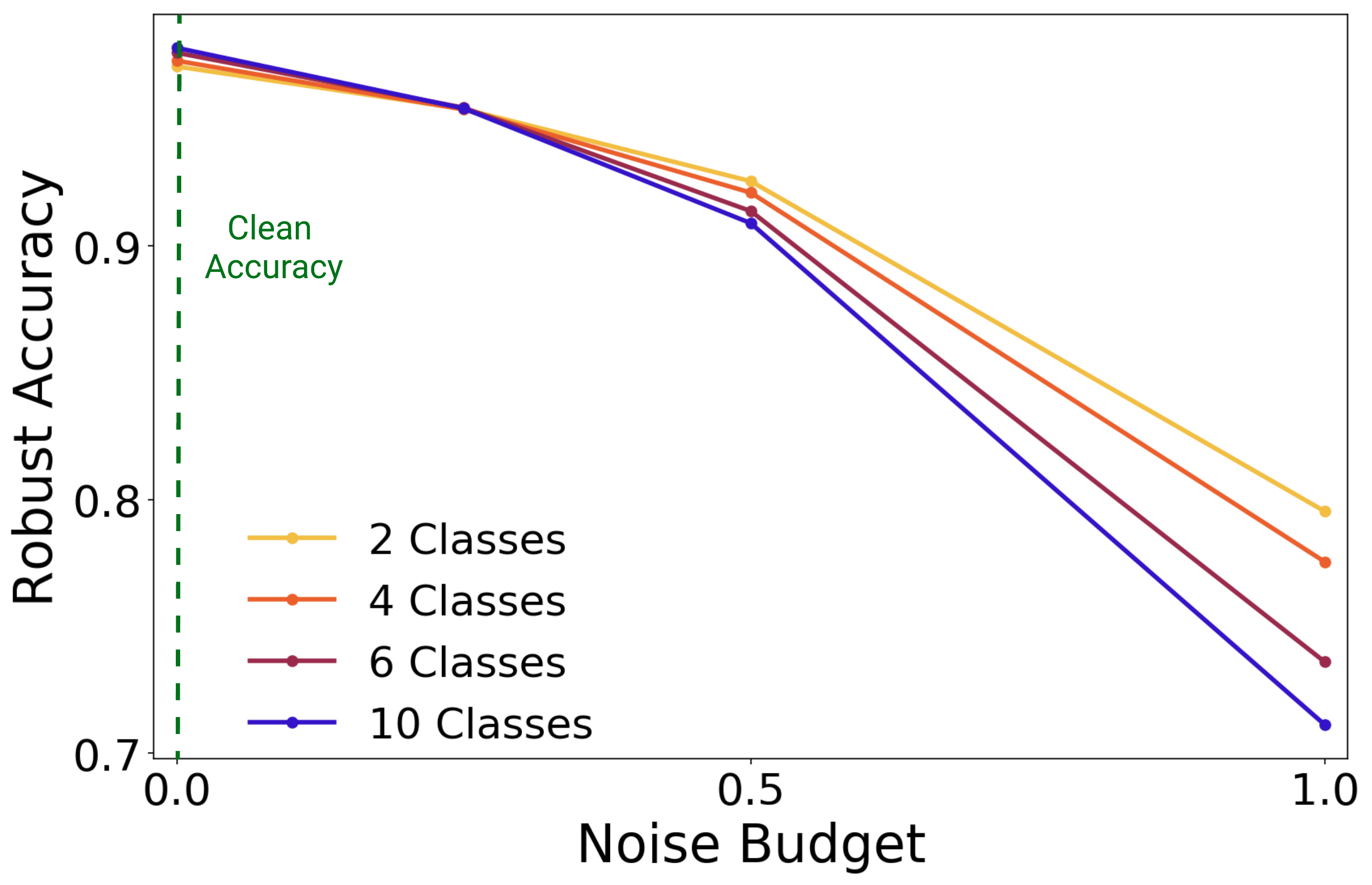}
    \caption{It is more difficult to attack simpler classifiers. Each line represents a different classifier trained over $k$ classes. Clean accuracy for all models is computed on the binary task. When the permitted test budget is the same as the one we train on ($\epsilon = 0.5$), simpler classes have higher performance. The advantage in robustness increases for higher values of $\epsilon$. We report the mean  over 20 runs of ResNet-18 on CIFAR-10.}
    \label{fig:less_better}
\end{minipage}
 \hspace{0.15cm} 
 \begin{minipage}[t]{0.48\linewidth}
    \centering
    \includegraphics[width=0.93\textwidth]{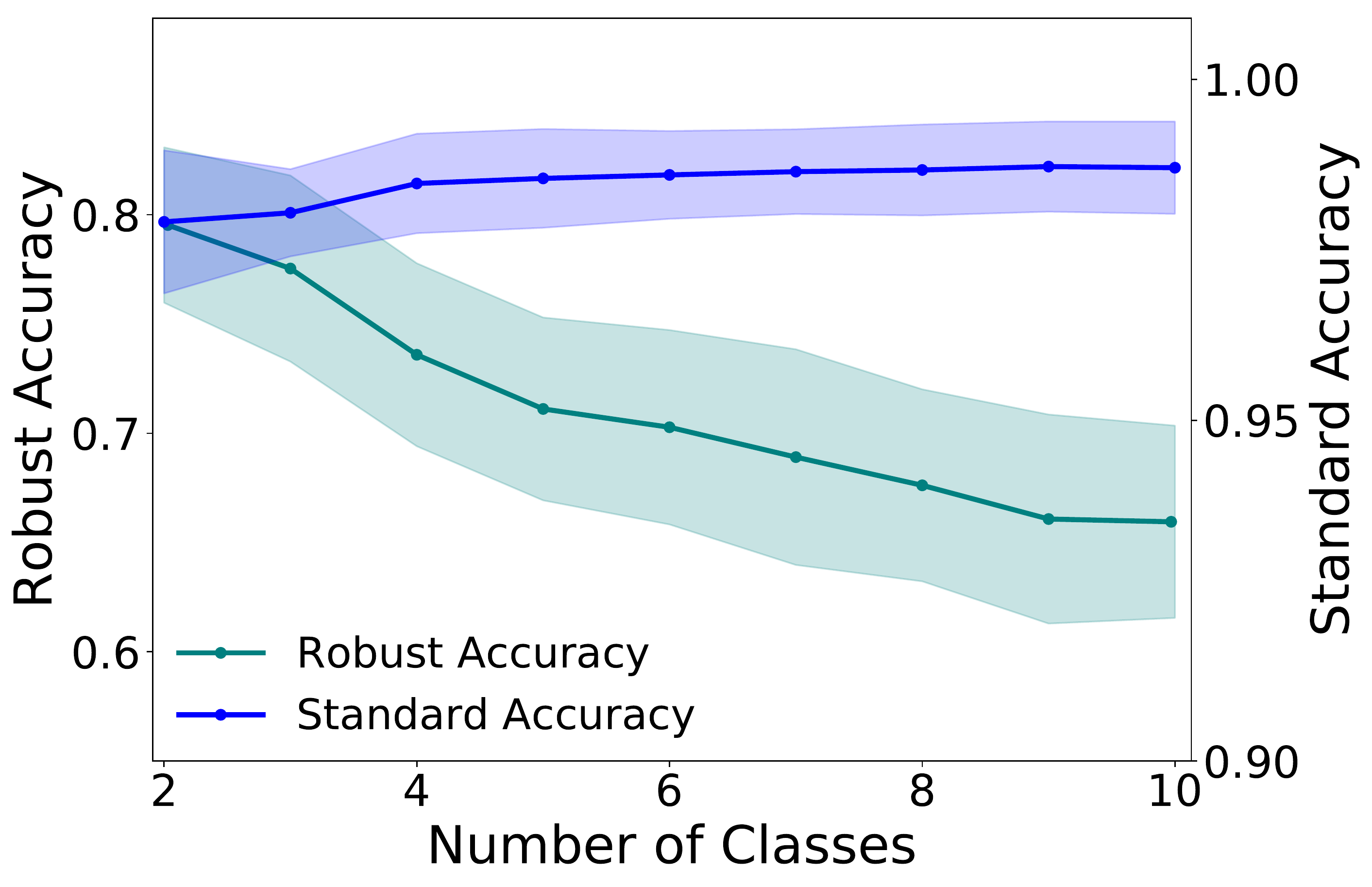}
    \caption{Difference in behavior between learning robust classifiers (teal, left y-axis) and learning non-robust classifiers for standard classification (blue, right y-axis). In the adversarial setting, increasing the number of output classes degrades performance (computed for $\epsilon = 1$), as opposed to the standard setting where accuracy slightly improves. We report mean and standard deviation over 20 class orderings of CIFAR-10 with ResNet-18.}
    \label{fig:subset_rob_eps} 
\end{minipage}
\end{center}
\end{figure*}

In this section, we investigate the relationship between robustness and model complexity by computing the average perturbation needed for samples to cross the decision boundary. Our main finding is that the robustness of a classifier decreases with the complexity of the task it performs. We first discuss a motivational example using Figure~\ref{fig:no_noise}, which shows learned decision boundaries of binary and multiclass models, before moving to experimental results.
\subsection{Motivational Example}
We want to compare the performance of two classifiers that are given the same task at test time: classify the presence of a dog in the image. The two classifiers are adversarially trained on the same data distribution, but differ solely on the information they are given at train time: The first binary classifier (Figure \ref{fig:no_noise}a) is given coarse labels, e.g., 1 if the picture contains a dog (blue) and 0 if it contains a cat or an airplane (red). The second, more complex classifier (Figure \ref{fig:no_noise}b) is given fine labels, e.g., 0 if the picture contains a dog (blue), 1 if the picture contains a cat (red) and 2 if the picture contains an airplane (green). Our goal is to understand which model will be more robust at test time.

When measuring robustness, we would like to control for varying complexity among models. Standard methods of computing robustness are shown for a binary classifier (Figure \ref{fig:no_noise}a) and a multiclass classifier (Figure \ref{fig:no_noise}b), where samples can be perturbed to the nearest decision boundary using untargeted attacks. Comparing these two figures shows that the potential increase in robustness in the binary classifier is purely gained from fewer decision boundaries. By only considering the decision boundary between dogs and other classes, we can compare the complex classifier to the binary one (Figure \ref{fig:no_noise}c). That is, we only consider a perturbation successful if either a dog is perturbed (untargeted attack) or if a cat or airplane is perturbed to be a dog (targeted attack).

\subsection{Empirical Results}

\paragraph{Experimental Setup.}
We first explain how to construct tasks of varying complexity. Assume we have a classification task with $k$ classes and we want to solve the binary one-vs-all problem for class 1. For each $j\in \{2,...,k\}$, define \texttt{OTHER[$j$]} to be a superclass containing classes $j,...,k$ and the corresponding \texttt{MODEL[$j$]} to be trained on the data with labels $\{1,2,3,...,j-1, $\texttt{ OTHER}$[j]\}$. Each \texttt{MODEL[$j$]} is trained on exactly $j$ distinct labels. We train a total of $k-1$ models, $\{\texttt{MODEL[$j$]}\}_{j=2}^k$, where \texttt{MODEL[2]} is the binary task and \texttt{MODEL[$k$]} is the full multiclass problem.

We measure robustness by considering a perturbation successful if either an example in class $1$ is perturbed to another class in $\{2,...,k\}$ (untargeted attack) or an example in class $j \in \{2,...,k\}$ is perturbed to class $1$ (targeted attack).

We repeat this experiment 20 times on CIFAR-10 with random permutations of class orderings to account for varying difficulty of the chosen target class (e.g., cat is harder to classify and easier to attack than airplane). We report the mean robustness accuracy during test time across models of fixed complexity for varying noise budgets $\epsilon$ in Figure~\ref{fig:less_better} and average perturbation sizes in Figure~\ref{fig:intro}. We defer further details on the training procedure to Appendix~\ref{app:simprob}.

\vspace{-.2cm}

\paragraph{Results.}
As seen in Figure \ref{fig:less_better}, there is a significant decrease in robust accuracy as the number of classes increases. 
Interestingly, reducing the number of output classes from 10 to 6 does not yield much robustness increase, but a further reduction to 4 or 2 classes makes the model significantly more robust. We note that even when the standard accuracy of the simpler classifier is slightly lower, the model is able to compensate with a gain in robustness.

\subsection{Separation of Learning Regimes} 
We first discuss the nuances in our previous observations. Then we address the qualitative difference in behaviors between standard and adversarial regimes. To measure standard accuracy, we repeat our experimental setup, but train classifiers over clean examples. Similar to the robust regime, at test time we evaluate all models on the binary task.

\vspace{-.2cm}
\paragraph{Conflicting Intuitions.} At first glance, there are two contradicting intuitions regarding the relationship between performance and complexity. On one hand, a simple classifier solves an easier task, i.e., if the complex classifier achieves perfect accuracy on the fine labels it trivially reduces to perfect performance on the coarse labels. Moreover, a simple classifier directly optimizes the test time objective. On the other hand, a complex classifier is given access to additional information on the data labels during training which could potentially increase performance.

\paragraph{Standard vs Robust.} Surprisingly, we observe different trends in standard and robust regimes. While robust learning coincides with our first intuition, as we previously discussed, we find that standard learning coincides with our second one. Specifically, the performance of the complex classifier is comparable to the simpler one and even slightly improves as a function of the number of output classes (Figure \ref{fig:subset_rob_eps}). However, in the robust setting, we see that simpler classifiers have a significant advantage over more complex ones.

We note that this disconnect between standard and adversarial settings is a testimony to the fundamental differences between the two problems. This is particularly intriguing as the adversarial setting can simply be viewed as optimizing a different (adversarial) loss function. Another such separation can be seen in the generalization behavior: In both settings, we can train models to have close to zero train error. However, in standard supervised learning, DNNs are able to generalize well with few examples while in the adversarial setting, their generalization is marginal \cite{madry2017towards}.

\section{Robust Aggregation of Binary Classifiers}\label{sec:algo}

We now leverage results from Section~\ref{sec:simprob} and combine simple classifiers for multiclass prediction. We show that \robin (RObust BINary Aggregation) achieves higher robustness than models trained directly on the multiclass task.

\subsection{Algorithm}

We decompose the original multiclass classification task into a set of one-vs-all binary tasks. Subsequently, we train binary classifiers and aggregate their outputs to perform multiclass prediction. We often refer to these binary classifiers as \emph{arms} of the aggregation.
Algorithm \ref{alg:pred} shows the general procedure where the original classes are transformed into binary labels and $k$ binary classifiers are adversarially trained. The predicted class of \robin is chosen by the binary classifier with the highest prediction score. While one-vs-all methods are a common approach (e.g., in SVMs), their usage in the context of robustness is largely unexplored.

\begin{algorithm}[t]
\caption{\textsc{ro}bust \textsc{bin}ary Aggregation (\textsc{robin})}
\label{alg:pred}
    \begin{algorithmic}[1]
     \STATE {\bf Input} data, ($\mathcal X$, $\mathcal Y$), number of epochs $T$, 
     \\step sizes $\eta_t, \hat {\eta}$, attack budget $\epsilon$.
      \STATE{$k \leftarrow$ number of classes in $\mathcal Y$}
      \STATE{$H \leftarrow \{ \}$  (set of binary classifiers)}
	\FOR{$i$ in $(1,...,k)$}
      \STATE {$\mathcal Y_i \leftarrow (\mathbbm{1}_{y_1 = i},...\mathbbm{1}_{y_m = i})$}
        \FOR{$t$ in $(1,...,T)$}
        \STATE{Generate adversarial example $\bx'$ using PGD with attack budget $\epsilon$ and step size $\hat \eta$.}
	        \STATE{Update parameters $\theta_t$}
	        \STATE{$\theta_{t+1} = \theta_t - \eta_t \nabla_{\theta_t} \cL(y, h_{\theta_t}(\bx')) $}
	\ENDFOR
       \STATE{Add $h^i = h_{\theta_T}$ to $H$}
     	\ENDFOR
      \STATE {\bf return} $\text{argmax}_{i\in |H|} h^i$
    \end{algorithmic}
 \end{algorithm}

\subsection{Attack Methods}
\label{sec:attacks}
While there are well-known attack methods for standalone DNNs (as described in Section~\ref{subsec:attack-alg}) as well as ensembles of multiclass classifiers \cite{tramer_ens}, these techniques do not directly translate to attacks for \robin. In this section, we discuss our most successful methods to attack \robin. Additional methods are discussed in Appendix~\ref{app:supp_res}. Any of the following methods can employ any attack algorithm (PGD, CW, etc.).

\begin{itemize*}
\setlength\itemsep{0.005em}
\item {\bf \textsc{Best Arm}.} We construct an adversarial example for each of the binary classifiers of the aggregation and consider the attack successful if any one of the constructed examples forces \robin to misclassify.

\item {\bf \textsc{Top 2}.} We attack the arms with the highest and the second-highest prediction scores simultaneously. We aim to decrease the score of the highest arm while increasing the score of the second highest one. For a gradient-based attack, this is achieved by stepping in the direction of the average of their gradients. 

\item {\bf \textsc{Softmax}.} For this attack, we view the aggregation as a single multiclass model by adding a \textsc{Softmax} layer over the predictions of the different binary classifiers. Since $\textsc{Softmax}$ is a continuous and differentiable relaxation of our prediction function, $\arg\max$, this modification provides a tractable way to attack \robin with gradient-based methods. This addresses the issue of obfuscated gradients ~\cite{obfuscated}, showing that the increase in robustness is not caused by the difficulty in computing gradients.

\end{itemize*}

We also consider both targeted and untargeted attacks transferred from the multiclass model. We find that crafting adversarial examples with the multiclass model as a surrogate is an ineffective attack. See more details in Appendix~\ref{app:transfer}.
In the following section, we use these attacks to evaluate the robustness of binary aggregations and compare the results to the analogous attacks on the single multiclass classifier.

\section{Experiments} \label{sec:exp}
\begin{figure}[t]
\begin{center}
  \includegraphics[width=.45\textwidth]{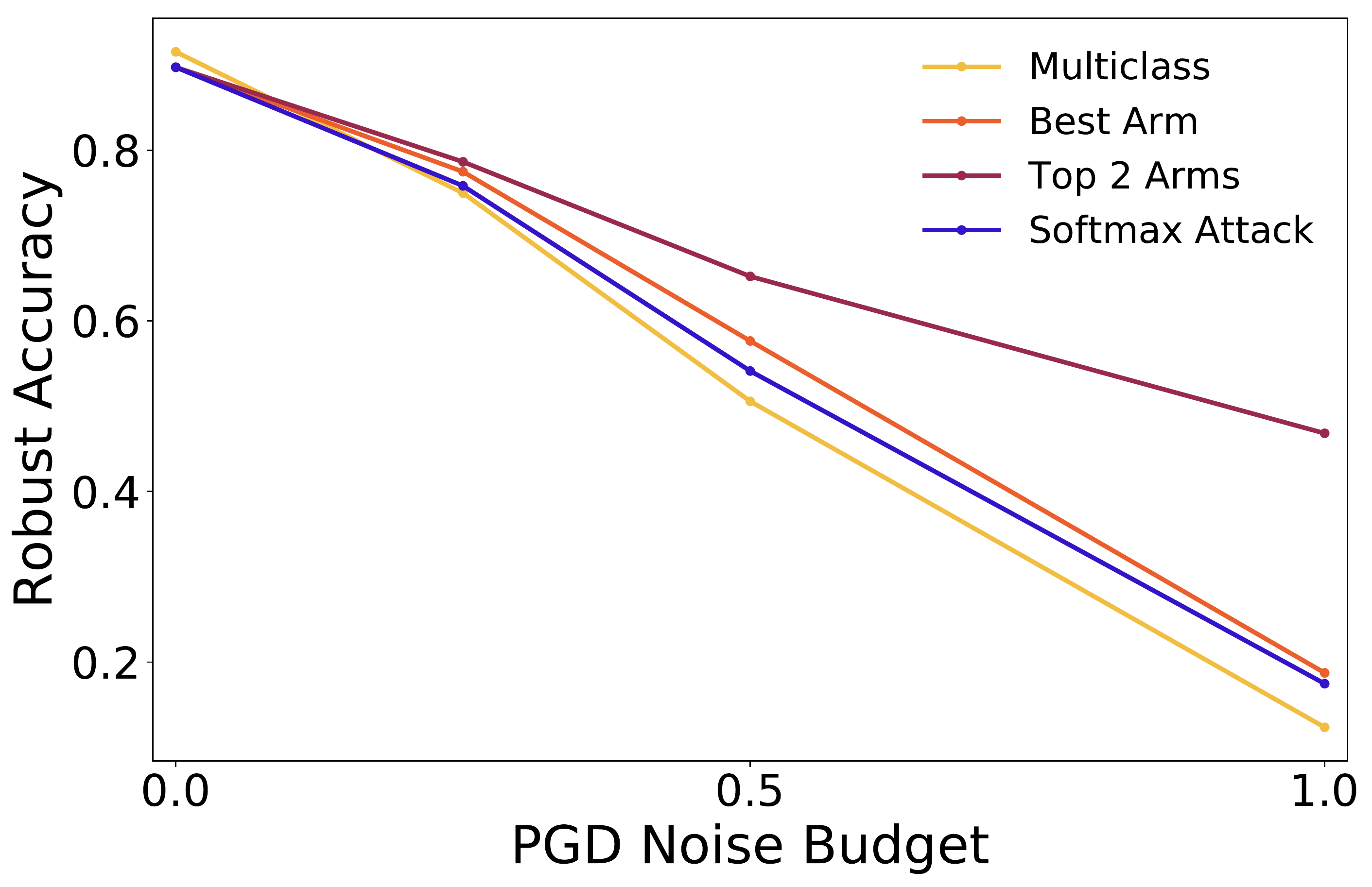}
\end{center}
\caption{Performance of three attacks on \robin and standard attack on the multiclass. Across all budgets, \textsc{Best Arm} and \textsc{Softmax} have comparable performance while \textsc{Top 2} is weaker. At the test budget of $\epsilon=0.5$, \robin has improved robustness over multiclass for all three attacks.}
\label{fig:attacks}
\end{figure}
In each set of experiments, we train $k$ binary classifiers and a single multiclass classifier to compare the robustness between \robin and the standard multiclass model. We use the same hyperparameters for both \robin and the multiclass. We now provide a description of our experimental setup. More details can be found in Appendix~\ref{app:setup}.

\begin{table*}[!ht]
\caption{Comparison of robust accuracy using different perturbation types between a standard multiclass classifier (MC) and \robin on CIFAR-10 and MNIST for various architectures. The $\ell_2$ noise budget allocated to the adversary (and used during training) is $\epsilon=0.5$ for CIFAR-10 and $\epsilon=2.0$ for MNIST. The respective $\ell_\infty$ budgets are $\epsilon = 8/255$ on CIFAR-10 and $\epsilon = 25/255$ on MNIST. }\label{tab:main-res}
\begin{center}
\begin{tabular}{c|c|c|cc|cc|cc|cc}
\toprule
\multirow{2}{*}{Dataset} & \multirow{2}{*}{Architecture}  & \multirow{2}{*}{Norm}  & 
\multicolumn{2}{c|}{Clean}
&
\multicolumn{2}{c|}{PGD10}
&
\multicolumn{2}{c|}{PGD20}

& \multicolumn{2}{c}{CW} \\                
 &  &  & MC & \robin & MC & \robin & MC & \robin & MC & \robin\\
\midrule
\multirow{6}{*}{CIFAR-10}
& \multirow{2}{*}{CNN}
& $\ell_2$ & 0.743  & \textbf{0.758} & 0.475 &  \textbf{0.555}& 0.474    &  \textbf{0.554} & 0.479 & \textbf{0.656}  \\
& & $\ell_\infty$  & \textbf{0.633} & 0.585 & 0.306 & \textbf{0.338} & 0.303 & \textbf{0.336} & 0.435 & \textbf{0.507}  \\
\cline{2-11}
& \multirow{2}{*}{ResNet-18} & $\ell_2$ & \textbf{0.890} & 0.869  & 0.617& \textbf{0.636} & 0.615 & \textbf{0.635} & 0.653 & \textbf{0.696} \\
& & $\ell_\infty$ & 0.766 & \textbf{0.805}  & 0.437 & \textbf{0.448} & 0.432 & \textbf{0.444} & 0.511 & \textbf{0.638}  \\
\cline{2-11}
& \multirow{2}{*}{ResNet-50} & $\ell_2$&  \textbf{0.906} & 0.883 & 0.614 & \textbf{0.637} & 0.612 & \textbf{0.634} & 0.659 & \textbf{0.705}\\
& & $\ell_\infty$ & 0.807 & \textbf{0.847}  & 0.444 & \textbf{0.461} & 0.439 & \textbf{0.447} & \textbf{0.620} & 0.614  \\
\midrule
\multirow{2}{*}{MNIST} & \multirow{2}{*}{CNN} & $\ell_2$ & \textbf{0.981} & 0.975  & 0.730 & \textbf{0.747} & 0.712 & \textbf{0.731} & 0.904 & \textbf{0.934} \\
& & $\ell_\infty$ & \textbf{0.992} & 0.991  & 0.958 & \textbf{0.959} & 0.957 & \textbf{0.959} & 0.986 & \textbf{0.988}  \\

\bottomrule
\end{tabular}
\end{center}
\end{table*}

\paragraph{Datasets and Models.} We evaluate the robustness of our models on the CIFAR-10 \cite{cifar} and MNIST \cite{mnist} datasets. For CIFAR-10 we use Myrtle CNN \cite{myrtle}, ResNet-18 and ResNet-50 \cite{res18}, while on MNIST we use a 2-layer CNN, CNN-2.

\paragraph{Defense Methods.} We train \robin and the multiclass baseline using three different defense methods.

\begin{itemize*}
\setlength\itemsep{0.05em}
\item {\bf Adversarial training.} 
In most experiments we use adversarial training, described in Section~\ref{sec:obj}, to train the binary classifiers and the baseline multiclass model.

\item {\bf\textsc{TRADES}.} 
Recently, \textsc{TRADES} \cite{trades} improves upon adversarial training by using a surrogate loss with an additional regularization term. Formally, the objective is:
\begin{align*}
    \min_\theta \E
    \left[\cL(h_\theta(\bx), y)
    + \lambda \textsc{KL}(h_\theta(\bx), h_\theta(\bx')) \right]
\end{align*}
 where \textsc{KL} denotes the KL-divergence. The first term corresponds to the natural loss and the second term regularizes the classifier to stay constant around $\bx$.
\item {\bf\textsc{MART}.} 
In a recent paper, \textsc{MART} \cite{mart} improves upon state-of-the-art robust accuracy by using a novel surrogate loss function. In particular, they optimize the following objective:
\begin{align*}
    \min_\theta \E \big[\textsc{BCE}(h_\theta(\bx), y) + &\\ \lambda \textsc{KL}(h_\theta(\bx), h_\theta(\bx'))&(1-h_\theta(\bx)_y) \big]
\end{align*}
where \textsc{BCE} is the Cross-Entropy loss with an additional margin term so that correct and misclassified examples are handled differently during training.

\end{itemize*}
We use default parameters of $\lambda =1$ in our implementation of TRADES and MART. With additional fine-tuning, performance for both could improve.
\paragraph{Training Procedure.}
We train all models with Stochastic Gradient Descent (SGD), mini-batches of size 128 and weight-decay of $5e$-$4$. For CIFAR-10, we train for 80 epochs for ResNet-18 and CNNs and 150 epochs for ResNet-50. For MNIST, we train for 20 epochs. We use an initial learning rate of 0.1 for the first half of training and decrease in magnitude for the remaining epochs except ResNet-50 where we decrease the learning rate every 50 epochs. For adversarial training, we use a warm-up schedule to gradually increase the proportion of adversarial examples in each batch, which is given by $1-\frac{1}{2}(1+\cos(\pi \times \frac{t}{T}))$ where $t$ is the current epoch and $T$ is the total number of epochs.  Adversarial examples are generated using 10 PGD steps. For CIFAR-10, we use random cropping with \texttt{padding}$=4$ for data-augmentation. Moreover, to account for class imbalance in one-vs-all binary classifiers, we sample positive and negative samples with similar rates. Details on the sampling process are deferred in Appendix~\ref{app:setup}.

\paragraph{Evaluation Procedure.}
We evaluate trained models on clean samples for standard accuracy and adversarially-perturbed samples for robust accuracy. The robustness at test time is computed with the same perturbation budget we used during training. We report the strongest of the three attacks from Section~\ref{sec:attacks} coupled with a base attack of PGD10, PGD20 (10 and 20 steps of PGD, respectively) and CW as the robustness of \robin. We use the same base attack to evaluate the multiclass model.

\subsection{Results}

In Table~\ref{tab:main-res} we present our experimental results for $\ell_2$ and $\ell_\infty$ with $\epsilon$ budgets of $0.5$ and $8/255$ respectively. We report the clean and robust accuracy for ResNet-18, ResNet-50 and Myrtle CNN for CIFAR-10 and CNN-2 for MNIST. In Table~\ref{tab:trades} we report the results for \textsc{TRADES} and MART.

Additionally, in Figure~\ref{fig:attacks}, we show the behavior of the different attack methods presented in Section~\ref{sec:attacks} as a function of the noise budget $\epsilon$ at test time. 

\paragraph{Increase in Robustness.} In our experiments, \robin consistently outperforms the multiclass model in robustness across different datasets, perturbation norms and model architectures. For a given $\epsilon$ budget, we see a 1-3\% increase in robustness using \robin for ResNets. The increase is significantly larger for CNNs on CIFAR-10. Interestingly, when a larger budget is used during test time compared to training, \robin becomes increasingly more robust than the multiclass (Figure~\ref{fig:attacks} and Appendix~\ref{app:supp_res}). In certain settings, \robin outperforms the benchmark in clean accuracy. 

Our results for \textsc{TRADES} and \textsc{MART} in Table~\ref{tab:trades} show that we still improve upon the robustness of the multiclass model when \robin and the benchmark are trained with these methods. For TRADES, we see a consistent 2\% increase in robustness. For MART, we see a slight improvement of 0.4\% in robustness as well as an improvement in clean accuracy over the benchmark.

As a corollary, we see that \robin is compatible with other training procedures and can be combined with future defense techniques. Furthermore, we note that \textsc{TRADES} and \textsc{MART} have loss functions specifically designed for the multiclass problem. We conjecture that a similar approach of crafting a loss function that directly optimizes for the binary problem can improve the results of \robin and leave this as an open problem for future work. 

\paragraph{Performance of Proposed Attacks.} While we compute the robustness of the standard model using one perturbation type, we evaluate \robin by considering the strongest of three different attacks discussed in Section~\ref{sec:attacks}. We find that the \textsc{Softmax} attack coupled with PGD is on average the strongest attack, with \textsc{Best Arm} achieving similar performance (Figure~\ref{fig:attacks}).

Since \textsc{Softmax}, a holistic attack on the composite classifier, is only marginally stronger than attacking a single arm, we can infer that the adversary does not gain much additional information when given access to multiple arms. Adversarial perturbations that fool one arm have limited ability to fool the others. In the following section, we attempt to formalize this claim by analyzing the gradient alignment of different arms of \robin, providing motivation to further explore similar techniques.

\begin{table}
\caption{Comparison of clean and robust accuracy between multiclass classifier/\robin for ResNet-18 on CIFAR-10. We trained both models using TRADES or MART. $\ell_2$ values correspond to $\epsilon=0.5$ and $\ell_\infty$ values correspond to $\epsilon = 8/255$.}\label{tab:trades}
\begin{center}
\begin{tabular}{c|c|c|c}
\toprule
Method & Norm & 
Clean
&
PGD10\\\hline
\multirow{2}{*}{TRADES}
& $\ell_2$ 
& \textbf{87.1} / 86.0  & 63.0 / \textbf{64.9}  \\
& $\ell_\infty$  
& \textbf{83.3} / 79.3  & 44.7 / \textbf{46.7}  \\
\cmidrule{1-4}
\multirow{2}{*}{MART}
& $\ell_2$ 
&  84.1 / \textbf{85.3}   & 62.7 / \textbf{63.0}   \\
& $\ell_\infty$  
& 75.5 / \textbf{76.1}  & 44.3 / \textbf{44.8}  \\
\bottomrule
\end{tabular}
\end{center}
\end{table}

\section{Gradient Alignment}\label{sec:gradient}

\begin{figure*}[t]
\begin{center}
  \includegraphics[width=.33\textwidth]{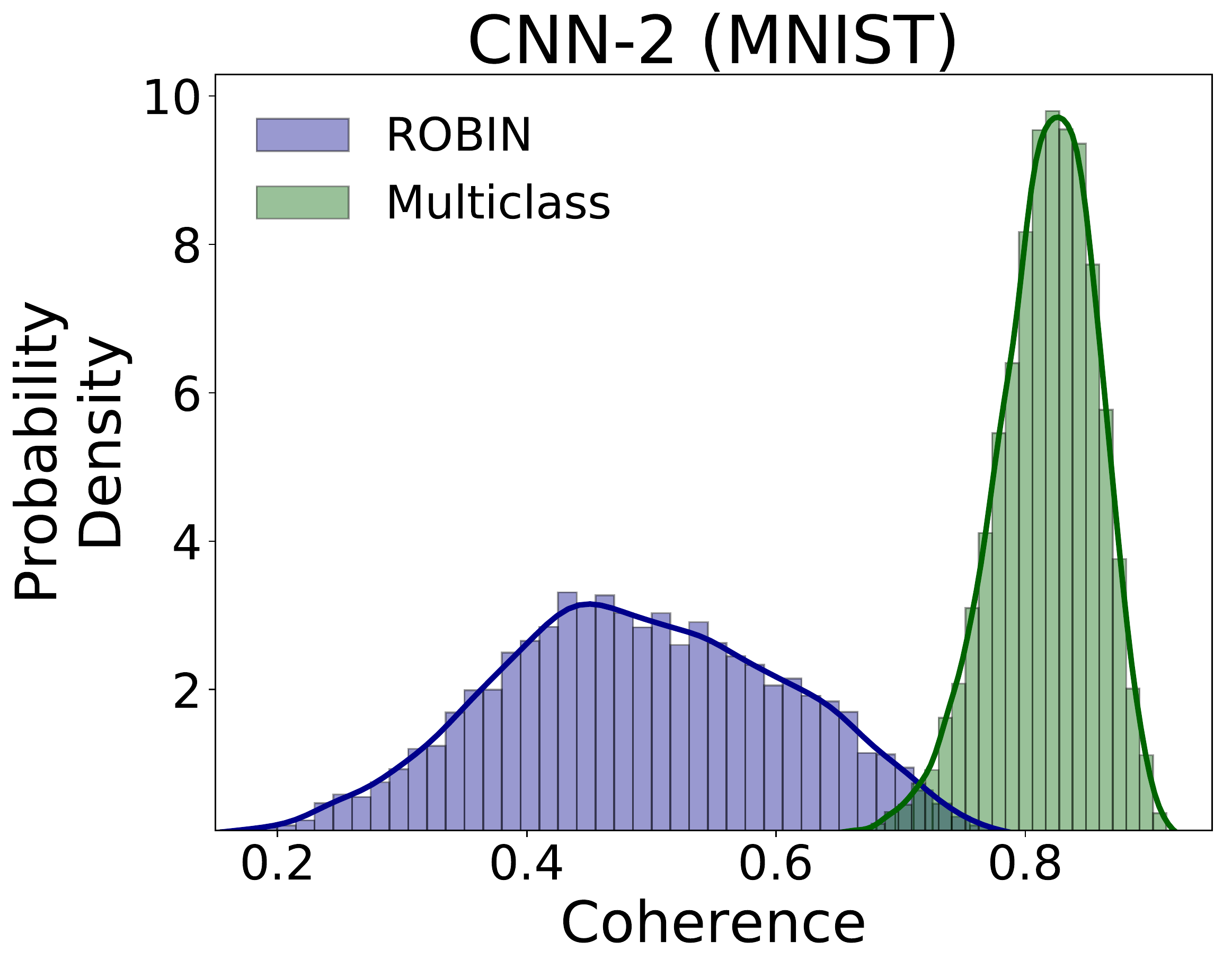}
  \includegraphics[width=.33\textwidth]{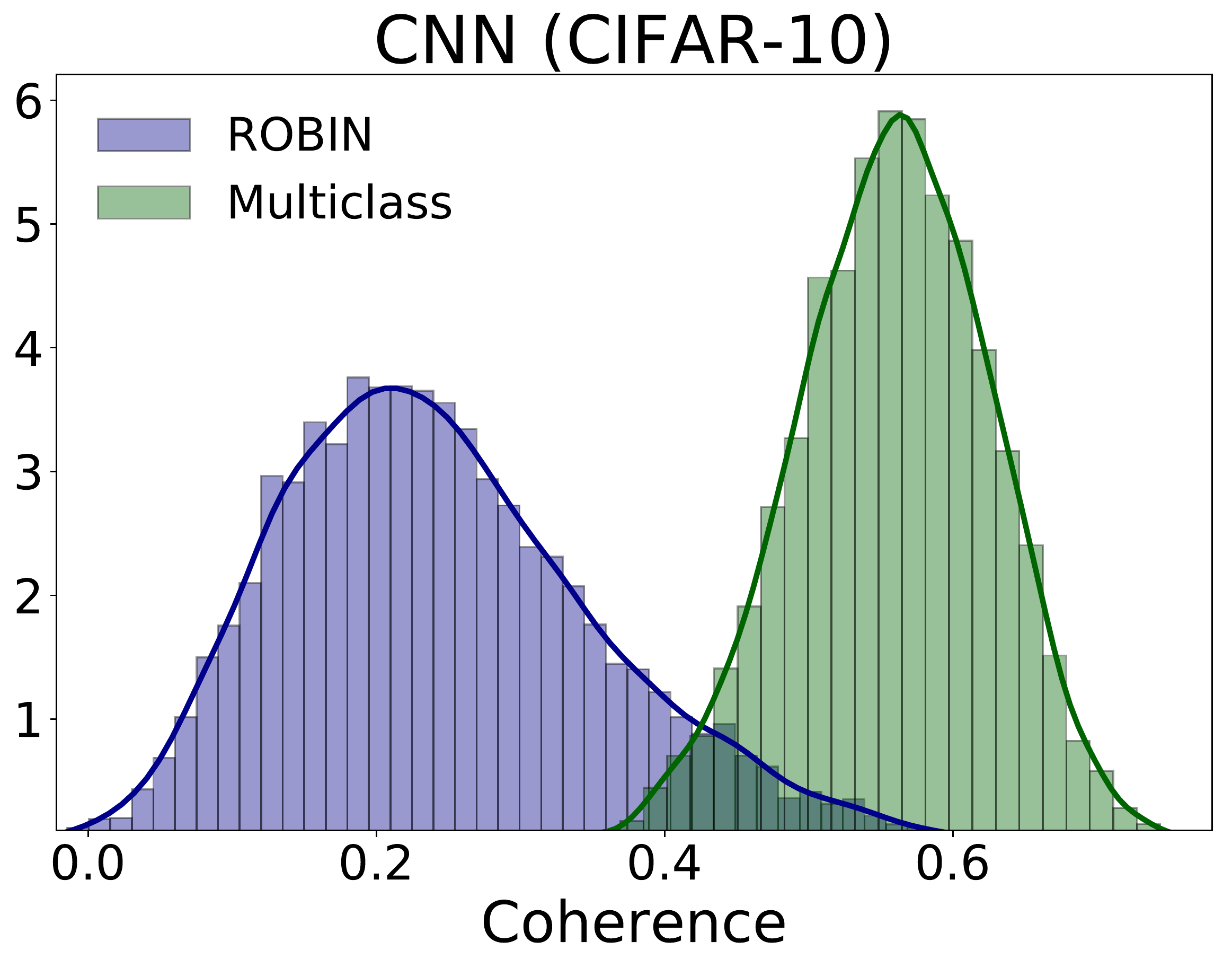}
  \includegraphics[width=.32\textwidth]{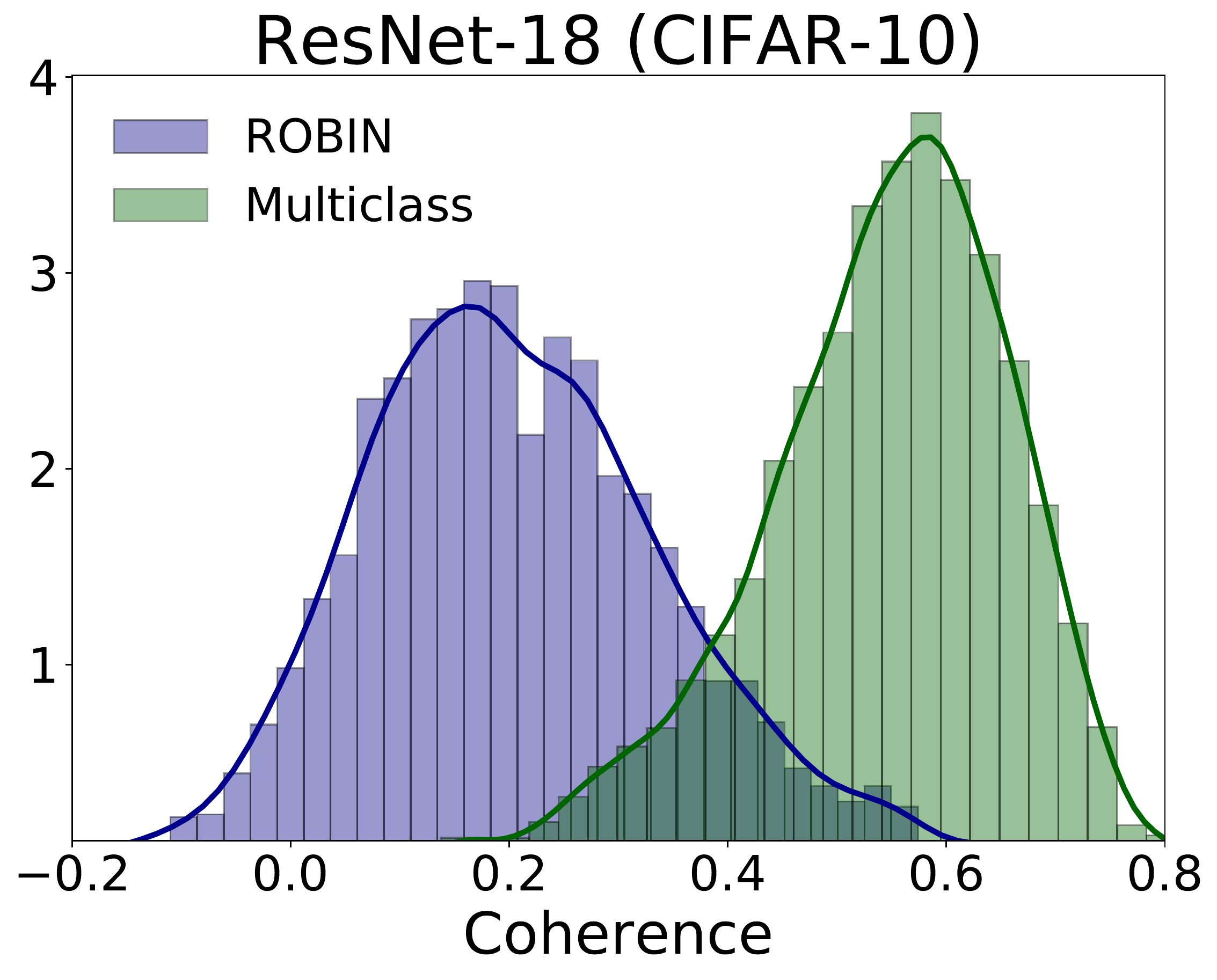}
\end{center}
\caption{Distribution of the coherence of \robin (purple) and a multiclass ensemble (green), across different datasets and architectures (CNN-2 on MNIST and Myrtle CNN, ResNet-18 on CIFAR-10). We compute the coherence of arms in \robin following Equation \ref{eqn:co} for each input. For the ensemble, we independently train 5 multiclass classifiers of the same architecture and compute the coherence over this set of models. The coherence of \robin is consistently shifted towards lower values.}
\label{fig:coherence}
\end{figure*}

In this section, we investigate the correlation between the gradients of the different arms of \robin. 
Through this analysis, we aim to provide an intuitive explanation for its success over the standard multiclass classifier.

In particular, to force \robin to misclassify an example, the adversary has to either completely manipulate one arm, or partially manipulate (at least) two arms. As we established earlier in Section \ref{sec:simprob}, manipulating one arm may be strictly harder than manipulating the multiclass model. On the other hand, deceiving two (or more) classifiers requires finding a distortion in the mutual adversarial space of two arms. Since the two gradients of these arms point in the direction of maximal loss for each individual classifier, we can use gradient alignment analysis to check whether it is likely that a direction is adversarial for both models.

\subsection{Coherence of Gradients}
Consider two models $h_0$ and $h_1$ and their respective gradients $\nabla\cL_0 := \nabla_\bx\cL(h_0(\bx),y)$ and $\nabla\cL_1 :=\nabla_\bx\cL(h_1(\bx),y)$ of the loss function with respect to the input $\bx$. Formally, the gradient describes the direction in which the input $\bx$ must be perturbed to maximally increase the linear approximation of the loss function around $\bx$. If $h_0$ and $h_1$ have gradients that are aligned, their loss functions have positive correlation: a perturbation that causes $\cL_0$ to increase will also cause $\cL_1$ to increase. This implies the two models share a common adversarial direction \cite{diversity-training}. Thus, from the perspective of the adversary, an aggregated model with aligned gradients is much easier to attack than a model with misaligned gradients.

We now formalize the measure of gradient alignment between two gradients using cosine similarity (CS):
\begin{eqnarray}
CS(\nabla \cL_0, \nabla\cL_1)) = \frac{\langle\nabla \cL_0, \nabla \cL_1\rangle}{|\nabla \cL_0| \cdot |\nabla \cL_1|}. \nonumber
\end{eqnarray}
To combine two models into a robust aggregation, ideally the cosine similarity of their gradients is close to $-1$. Then, the adversarial directions of the models are antiparallel and, consequently, the common adversarial space is small. Similarly, for $k$ models we would like the gradient vector of the arm corresponding to the true label to be misaligned with the gradients of the rest $k-1$ arms. Then, one cannot significantly decrease the score of the correct arm while simultaneously increasing the score of an incorrect one.

We define the gradient alignment across $k$ models using coherence  \cite{tropp2006}, which we compute by considering the cosine similarity between the gradient of the binary classifier corresponding to the true label and the gradients of all other arms. Formally,
\begin{eqnarray}\label{eqn:co}
coherence(\{\nabla_\bx \cL_i\}_{i=1}^k) = \max_{j\neq y_\bx} CS(\nabla_\bx \cL_{y_\bx}, \nabla_\bx \cL_j).
\end{eqnarray}
The coherence or maximum cosine similarity provides a measure of gradient alignment for each input.

\subsection{Coherence of \robin}
In Figure~\ref{fig:coherence}, we plot the distribution of coherence over gradients of \robin for each input and show that it is consistently low (e.g., about 0.2 average coherence for ResNet-18 on CIFAR-10) across different architectures and datasets. For reference, we also plot the coherence of a multiclass ensemble---a set of 5 networks (10 gradient pairs) trained independently, that collectively make predictions---and observe that the average coherence is much higher.  Hence, the adversarial directions of the arms are almost orthogonal, and consequently, an adversary would be mostly unable to significantly influence multiple arms.

As a result, attacking \robin with the \textsc{Softmax} attack (which attacks multiple arms simultaneously) is not significantly more effective than attacking each arm individually as in the \textsc{Best arm} attack. Additionally, we found that the examples \textsc{Softmax} successfully perturbs are mostly the same as the examples successfully perturbed by \textsc{Best arm}.
For example for ResNet-18 in CIFAR-10 with $\ell_2$ perturbations, \textsc{Best Arm} manages to successfully perturb 3263 examples, while \textsc{Softmax} 3641 examples. Out of these, 3259 examples are common among the two attacks. 
This provides further evidence to support the claim. 

\section{Discussion}
\label{sec:discussion}
We introduce a new notion of robustness from simplicity, i.e., models enjoy robust performance upon reduction in the number of classes. This contribution might be of independent interest since many practical applications of classification can be simplified by reducing the number of classes. For example, an airport security service might want to screen images for dangerous items. Instead of classifying hundreds of different items, the task can be simplified to a binary classification task of detecting a hazardous item in the image.

Additionally, the robust setting stands in stark contrast with the standard learning setting where \emph{adding} more classes can help learning by providing additional information about the labels. This is particularly intriguing as we have yet to understand the similarities and differences between standard and robust settings, specifically why models successfully generalize in the former and fail to do so in the latter. 

On a different note, while our method improves robustness over standard architectures, there is an increase in the number of model parameters, especially when the number of classes is large. However, we note that solely increasing the size of a model is usually insufficient to gain robustness. For example, Table~\ref{tab:main-res} shows that ResNet-18 has comparable robustness to ResNet-50. Moreover, multiple works propose ensembles of classifiers which increase the model capacity similarly to \robin. However, the most successful ensemble techniques use regularization that requires classifiers to not be trained independently, making training intractable for larger ensembles \cite{diversity-training}. We avoid the scalability issue as the arms of \robin can be trained independently and in parallel.

In addition to the one-vs-all decomposition, we tried other aggregation techniques to investigate different decompositions of the output space. In particular, we reduced the multiclass problem into a compilation of one-vs-one classifiers and into two-level hierarchical classifiers. Unfortunately, the first approach suffered a large clean accuracy drop while the second approach had poor robustness performance. Details on these experiments can be found in Appendix~\ref{app:stacked}. We leave more complex aggregations as an important avenue for future work.

\section{Conclusion}\label{sec:conclusion}

In this paper, we investigate the connection between robustness and simplicity and show an inverse relationship, i.e., it is increasingly difficult to perturb an example when decreasing the complexity of the classification task. Motivated by this observation, we propose \robin---a binary classifier aggregation method---to enhance robustness of state-of-the-art methods. We show that this method outperforms the multiclass benchmark in terms of robust generalization and has comparable accuracy on clean images. Moreover, we show that \robin is complementary to other training techniques and can be combined to further boost robust accuracy.
\section*{Acknowledgements}

We would like to thank Preetum Nakkiran and Boaz Barak for early discussions and suggestions on this work. We also thank Nick Roy for comments on early drafts of this work.

This work was supported by Yaron Singer's NSF grant CAREER CCF 1452961, NSF CCF 1301976, BSF grant 2014389, NSF USICCS proposal 1540428, Google Research award, Facebook research award, and Boaz Barak's Simons investigator fellowship and a gift from Oracle labs.

\bibliography{main}

\begin{thebibliography}{28}
\providecommand{\natexlab}[1]{#1}
\providecommand{\url}[1]{\texttt{#1}}
\expandafter\ifx\csname urlstyle\endcsname\relax
  \providecommand{\doi}[1]{doi: #1}\else
  \providecommand{\doi}{doi: \begingroup \urlstyle{rm}\Url}\fi

\bibitem[Athalye et~al.(2017)Athalye, Engstrom, Ilyas, and Kwok]{synthesizing}
Athalye, A., Engstrom, L., Ilyas, A., and Kwok, K.
\newblock Synthesizing robust adversarial examples.
\newblock \emph{arXiv preprint arXiv:1707.07397}, 2017.

\bibitem[Athalye et~al.(2018)Athalye, Carlini, and Wagner]{obfuscated}
Athalye, A., Carlini, N., and Wagner, D.~A.
\newblock Obfuscated gradients give a false sense of security: Circumventing
  defenses to adversarial examples.
\newblock \emph{CoRR}, abs/1802.00420, 2018.
\newblock URL \url{http://arxiv.org/abs/1802.00420}.

\bibitem[Carlini \& Wagner(2017)Carlini and Wagner]{cw}
Carlini, N. and Wagner, D.
\newblock Towards evaluating the robustness of neural networks.
\newblock In \emph{2017 ieee symposium on security and privacy (sp)}, pp.\
  39--57. IEEE, 2017.

\bibitem[Ding et~al.(2018)Ding, Sharma, Lui, and Huang]{mma}
Ding, G.~W., Sharma, Y., Lui, K. Y.~C., and Huang, R.
\newblock Max-margin adversarial {(MMA)} training: Direct input space margin
  maximization through adversarial training.
\newblock \emph{CoRR}, abs/1812.02637, 2018.
\newblock URL \url{http://arxiv.org/abs/1812.02637}.

\bibitem[Feinman et~al.(2017)Feinman, Curtin, Shintre, and Gardner]{detect1}
Feinman, R., Curtin, R.~R., Shintre, S., and Gardner, A.~B.
\newblock Detecting adversarial samples from artifacts.
\newblock \emph{CoRR}, abs/1703.00410, 2017.
\newblock URL \url{http://arxiv.org/abs/1703.00410}.

\bibitem[Goodfellow et~al.(2014)Goodfellow, Shlens, and
  Szegedy]{goodfellow2014explaining}
Goodfellow, I.~J., Shlens, J., and Szegedy, C.
\newblock Explaining and harnessing adversarial examples.
\newblock \emph{arXiv preprint arXiv:1412.6572}, 2014.

\bibitem[Goodfellow et~al.(2015)Goodfellow, Shlens, and Szegedy]{goodfellow}
Goodfellow, I.~J., Shlens, J., and Szegedy, C.
\newblock Explaining and harnessing adversarial examples.
\newblock In Bengio, Y. and LeCun, Y. (eds.), \emph{3rd International
  Conference on Learning Representations, {ICLR} 2015, San Diego, CA, USA, May
  7-9, 2015, Conference Track Proceedings}, 2015.

\bibitem[He et~al.(2016)He, Zhang, Ren, and Sun]{res18}
He, K., Zhang, X., Ren, S., and Sun, J.
\newblock Deep residual learning for image recognition.
\newblock In \emph{Proceedings of the IEEE conference on computer vision and
  pattern recognition}, pp.\  770--778, 2016.

\bibitem[He et~al.(2017)He, Wei, Chen, Carlini, and Song]{attackensembles}
He, W., Wei, J., Chen, X., Carlini, N., and Song, D.
\newblock Adversarial example defenses: Ensembles of weak defenses are not
  strong.
\newblock \emph{CoRR}, abs/1706.04701, 2017.
\newblock URL \url{http://arxiv.org/abs/1706.04701}.

\bibitem[Kariyappa \& Qureshi(2019)Kariyappa and Qureshi]{diversity-training}
Kariyappa, S. and Qureshi, M.~K.
\newblock Improving adversarial robustness of ensembles with diversity
  training.
\newblock \emph{arXiv preprint arXiv:1901.09981}, 2019.

\bibitem[Krizhevsky(2009)]{cifar}
Krizhevsky, A.
\newblock Learning multiple layers of features from tiny images.
\newblock \emph{CoRR}, 2009.

\bibitem[Kurakin et~al.(2016)Kurakin, Goodfellow, and Bengio]{bim}
Kurakin, A., Goodfellow, I., and Bengio, S.
\newblock Adversarial examples in the physical world.
\newblock \emph{arXiv preprint arXiv:1607.02533}, 2016.

\bibitem[LeCun et~al.(1998)LeCun, Bottou, Bengio, and Haffner]{mnist}
LeCun, Y., Bottou, L., Bengio, Y., and Haffner, P.
\newblock Gradient-based learning applied to document recognition.
\newblock \emph{Proceedings of the IEEE}, 86\penalty0 (11):\penalty0
  2278--2324, 1998.

\bibitem[Liu et~al.(2019)Liu, Liu, Liu, Xu, Lin, Wang, and Wen]{compression}
Liu, Z., Liu, Q., Liu, T., Xu, N., Lin, X., Wang, Y., and Wen, W.
\newblock Feature distillation: Dnn-oriented {JPEG} compression against
  adversarial examples.
\newblock In \emph{{IEEE} Conference on Computer Vision and Pattern
  Recognition, {CVPR} 2019, Long Beach, CA, USA, June 16-20, 2019}, pp.\
  860--868. Computer Vision Foundation / {IEEE}, 2019.

\bibitem[Ma et~al.(2018)Ma, Li, Wang, Erfani, Wijewickrema, Schoenebeck, Song,
  Houle, and Bailey]{detect18}
Ma, X., Li, B., Wang, Y., Erfani, S.~M., Wijewickrema, S. N.~R., Schoenebeck,
  G., Song, D., Houle, M.~E., and Bailey, J.
\newblock Characterizing adversarial subspaces using local intrinsic
  dimensionality.
\newblock In \emph{6th International Conference on Learning Representations,
  {ICLR} 2018, Vancouver, BC, Canada, April 30 - May 3, 2018, Conference Track
  Proceedings}, 2018.

\bibitem[Madry et~al.(2017)Madry, Makelov, Schmidt, Tsipras, and
  Vladu]{madry2017towards}
Madry, A., Makelov, A., Schmidt, L., Tsipras, D., and Vladu, A.
\newblock Towards deep learning models resistant to adversarial attacks.
\newblock \emph{arXiv preprint arXiv:1706.06083}, 2017.

\bibitem[Page(2018)]{myrtle}
Page, D.
\newblock How to train your resnet.
\newblock In \emph{https://myrtle.ai/how-to-train-your-resnet-4-architecture/},
  2018.

\bibitem[Pang et~al.(2018)Pang, Du, Dong, and Zhu]{detect2}
Pang, T., Du, C., Dong, Y., and Zhu, J.
\newblock Towards robust detection of adversarial examples.
\newblock In Bengio, S., Wallach, H.~M., Larochelle, H., Grauman, K.,
  Cesa{-}Bianchi, N., and Garnett, R. (eds.), \emph{Advances in Neural
  Information Processing Systems 31: Annual Conference on Neural Information
  Processing Systems 2018, NeurIPS 2018, 3-8 December 2018, Montr{\'{e}}al,
  Canada}, pp.\  4584--4594, 2018.

\bibitem[Pang et~al.(2019)Pang, Xu, Du, Chen, and Zhu]{promo-diversity}
Pang, T., Xu, K., Du, C., Chen, N., and Zhu, J.
\newblock Improving adversarial robustness via promoting ensemble diversity.
\newblock \emph{arXiv preprint arXiv:1901.08846}, 2019.

\bibitem[Papernot et~al.(2016)Papernot, McDaniel, Wu, Jha, and
  Swami]{def_distil}
Papernot, N., McDaniel, P.~D., Wu, X., Jha, S., and Swami, A.
\newblock Distillation as a defense to adversarial perturbations against deep
  neural networks.
\newblock In \emph{{IEEE} Symposium on Security and Privacy, {SP} 2016, San
  Jose, CA, USA, May 22-26, 2016}, pp.\  582--597. {IEEE} Computer Society,
  2016.

\bibitem[Papernot et~al.(2017)Papernot, McDaniel, Goodfellow, Jha, Celik, and
  Swami]{masking}
Papernot, N., McDaniel, P.~D., Goodfellow, I.~J., Jha, S., Celik, Z.~B., and
  Swami, A.
\newblock Practical black-box attacks against machine learning.
\newblock In \emph{Proceedings of the 2017 {ACM} on Asia Conference on Computer
  and Communications Security, AsiaCCS 2017, Abu Dhabi, United Arab Emirates,
  April 2-6, 2017}, pp.\  506--519. {ACM}, 2017.

\bibitem[Ross \& Doshi{-}Velez(2018)Ross and Doshi{-}Velez]{RDV18}
Ross, A.~S. and Doshi{-}Velez, F.
\newblock Improving the adversarial robustness and interpretability of deep
  neural networks by regularizing their input gradients.
\newblock In \emph{Proceedings of the Thirty-Second {AAAI} Conference on
  Artificial Intelligence, (AAAI-18), the 30th innovative Applications of
  Artificial Intelligence (IAAI-18), and the 8th {AAAI} Symposium on
  Educational Advances in Artificial Intelligence (EAAI-18), New Orleans,
  Louisiana, USA, February 2-7, 2018}, pp.\  1660--1669. {AAAI} Press, 2018.

\bibitem[Szegedy et~al.(2013)Szegedy, Zaremba, Sutskever, Bruna, Erhan,
  Goodfellow, and Fergus]{szegedy2013intriguing}
Szegedy, C., Zaremba, W., Sutskever, I., Bruna, J., Erhan, D., Goodfellow, I.,
  and Fergus, R.
\newblock Intriguing properties of neural networks.
\newblock \emph{arXiv preprint arXiv:1312.6199}, 2013.

\bibitem[Tram{\`{e}}r et~al.(2018)Tram{\`{e}}r, Kurakin, Papernot, Goodfellow,
  Boneh, and McDaniel]{tramer_ens}
Tram{\`{e}}r, F., Kurakin, A., Papernot, N., Goodfellow, I.~J., Boneh, D., and
  McDaniel, P.~D.
\newblock Ensemble adversarial training: Attacks and defenses.
\newblock In \emph{6th International Conference on Learning Representations,
  {ICLR} 2018, Vancouver, BC, Canada, April 30 - May 3, 2018, Conference Track
  Proceedings}, 2018.

\bibitem[Tropp et~al.(2006)Tropp, Gilbert, and Strauss]{tropp2006}
Tropp, J.~A., Gilbert, A.~C., and Strauss, M.~J.
\newblock Algorithms for simultaneous sparse approximation. part i: Greedy
  pursuit.
\newblock \emph{Signal processing}, 86\penalty0 (3):\penalty0 572--588, 2006.

\bibitem[Wang et~al.(2020)Wang, Zou, Yi, Bailey, Ma, and Gu]{mart}
Wang, Y., Zou, D., Yi, J., Bailey, J., Ma, X., and Gu, Q.
\newblock Improving adversarial robustness requires revisiting misclassified
  examples.
\newblock In \emph{International Conference on Learning Representations}, 2020.
\newblock URL \url{https://openreview.net/forum?id=rklOg6EFwS}.

\bibitem[Xu et~al.(2018)Xu, Evans, and Qi]{feat_sq}
Xu, W., Evans, D., and Qi, Y.
\newblock Feature squeezing: Detecting adversarial examples in deep neural
  networks.
\newblock In \emph{25th Annual Network and Distributed System Security
  Symposium, {NDSS} 2018, San Diego, California, USA, February 18-21, 2018}.
  The Internet Society, 2018.

\bibitem[Zhang et~al.(2019)Zhang, Yu, Jiao, Xing, Ghaoui, and Jordan]{trades}
Zhang, H., Yu, Y., Jiao, J., Xing, E.~P., Ghaoui, L.~E., and Jordan, M.~I.
\newblock Theoretically principled trade-off between robustness and accuracy.
\newblock \emph{arXiv preprint arXiv:1901.08573}, 2019.

\end{thebibliography}
\bibliographystyle{icml2020}

\newpage
\appendix
\begin{table*}[!ht]
  \caption{Performance of different attacks on \robin for varying $\epsilon$ at test time. \textsc{Softmax} is consistently the strongest attack with \textsc{Best Arm} as a comparable attack in some settings.}
\label{tab:attacks}
 \centering
  \begin{tabular}{cc|cccccc}
  \toprule

& \multicolumn{7}{c}{ResNet-18 CIFAR-10}\\
 \toprule
 
 
& $\ell_2$ radius  &  \multicolumn{6}{c}{Attacks (Train $\epsilon = 0.5$)}    \\ \hline
\multirow{5}{*}{\rotatebox[origin=c]{90}{Test $\epsilon$}} & &  \textsc{Best Arm} & \textsc{Top 2} & \textsc{Softmax} & \textsc{Highest Arm} & \textsc{Avg Grad} & \textsc{Softmax Top 2}\\
& $\epsilon = 0.25$  & 0.7811 & 0.7879 & \textbf{0.765}   &     0.7901 & 0.7878 & 0.8169              \\
& $\epsilon = 0.50$  & 0.6738 & 0.6937 & \textbf{0.636}   &     0.6978 & 0.6937 & 0.7081   \\
& $\epsilon = 0.75$ & 0.5319 & 0.5774 & \textbf{0.479}   &     0.5824 & 0.5773 & 0.56 \\

& $\epsilon = 1.00$  & 0.3843 & 0.4633 & \textbf{0.3395}   &     0.4539 & 0.4626 & 0.4275 \\

\midrule
& $\ell_\infty$ radius 
&  \multicolumn{6}{c}{Attacks (Train $\epsilon = 8/255$)} \\ 
\hline
\multirow{5}{*}{\rotatebox[origin=c]{90}{Test $\epsilon$}} & &  \textsc{Best Arm} & \textsc{Top 2} & \textsc{Softmax} & \textsc{Highest Arm} & \textsc{Avg Grad} & \textsc{Softmax Top 2} \\
& $\epsilon = 4/255$  & 0.6481 & 0.6583 & \textbf{0.6395}   &     0.6838 & 0.6677 & 0.7158                \\
& $\epsilon = 6/255$  & 0.558 & 0.5802 & \textbf{0.5454}   &     0.6133 & 0.591 & 0.632    \\
& $\epsilon = 8/255$  & 0.4666 & 0.4991 & \textbf{0.4484}   &     0.5319 & 0.5095 & 0.5436\\

& $\epsilon = 10/255$  & 0.3742 & 0.4206 & \textbf{0.3588}   &     0.4534 & 0.4302 & 0.4599\\
\bottomrule
\end{tabular}
 \end{table*}

\section{Experimental Details for Section \ref{sec:simprob}.}\label{app:simprob}
We train all models on CIFAR-10 using ResNet-18. The details below are similar to the experimental details from Section~\ref{sec:exp}. We use Stochastic Gradient Descent (SGD), mini-batches of size 128 and weight-decay of $5e$-$4$. We train for 80 epochs for ResNet-18 with an initial learning rate of 0.1 for the first half of training and decrease in magnitude for the remaining epochs. All models are trained using adversarial training with a warm-up schedule to gradually increase the proportion of adversarial examples in each batch, which is given by $1-\frac{1}{2}(1+\cos(\pi \times \frac{t}{80}))$ where $t$ is the current epoch. Adversarial examples are generated using 10 PGD steps with an $\ell_2$ norm budget of $\epsilon=0.5$. Additionally, we perform data augmentation, where we use random cropping with \texttt{padding}$=4$.

To evaluate robustness across models, we use the same noise budget at test time as we used in training.
\section{Supplementary Results for Section \ref{sec:exp}}\label{app:algo}
\subsection{Additional Attacks for \robin} \label{app:supp_res}

In addition to the attacks mentioned in Section~\ref{sec:attacks}, we considered the following three methods to attack \robin. As before, they can be combined with attacks like PGD or Carlini-Wagner.
\begin{itemize*}
\setlength\itemsep{0.05em}
    \item {\bf \textsc{Highest score Arm}.} One of the most natural attacks to consider is to use the entire noise budget on the arm with the highest prediction score. This attack will trivially be at most as strong as the \textsc{Best Arm} attack mentioned in Section~\ref{sec:attacks}.
    \item {\bf \textsc{Average Gradient}.} We attack all arms simultaneously and take a step in the direction of the average of their gradients. The goal of this attack is to decrease the prediction score for the correct arm while increasing the score of the rest of the arms.
    \item {\bf \textsc{Softmax top 2}.} This attack is a combination of the \textsc{TOP 2} and \textsc{Softmax} attacks described in Section~\ref{sec:attacks}. We attack the smooth approximation of the two arms with the highest prediction scores by isolating these two arms and applying a softmax layer on them.
\end{itemize*}

In Table~\ref{tab:attacks} we summarize the performance of the different attack methods we consider for a ResNet-18 on CIFAR-10. The model is adversarially trained using PGD10 with $\epsilon = 0.5$ for $\ell_2$ and $\epsilon = 8/255$ for $\ell_\infty$. A subset of this table was presented in Figure~\ref{fig:attacks} in Section~\ref{sec:exp}.

As we argue in the main body of the paper, the \textsc{Softmax} attack, presented in Section~\ref{sec:attacks}, is the most successful one. This is confirmed in Table~\ref{tab:attacks} which highlights the strongest attack in bold.

\subsection{Transferability from Multiclass} \label{app:transfer}

\begin{table}[h!]
\caption{Robust accuracy of \robin on adversarial examples transferred from the multiclass benchmark across varying $\epsilon$ during test time. For $\ell_2$, the models are trained with $\epsilon = 0.5$ while for $\ell_\infty$, models are trained with $\epsilon  = 8/255$.}
\label{tab:transferability}
\begin{center}
\begin{tabular}{cc|c|c}
  \toprule
\multicolumn{4}{c}{ResNet-18 CIFAR-10}\\
\toprule
& Radius & 
Untargeted
&
Targeted\\\hline
\multirow{4}{*}{\rotatebox[origin=c]{90}{$\ell_2$ test $\epsilon$}}
& $\epsilon = 0.25$ 
& 0.8285  & 0.8517  \\
& $\epsilon = 0.50$ 
& 0.7759  & 0.8182  \\
& $\epsilon = 0.75$ 
& 0.7179  & 0.767  \\
& $\epsilon = 1.00$ 
& 0.6441  & 0.7037  \\
\cmidrule{1-4}
\multirow{4}{*}{\rotatebox[origin=c]{90}{$\ell_\infty$ test $\epsilon$}}
& $\epsilon = 4/255$ 
&  0.7302   & 0.7819   \\
& $\epsilon=6/255$  
& 0.6838  & 0.7596  \\
& $\epsilon = 8/255$ 
& 0.6349  & 0.7261 \\
& $\epsilon = 10/255$ 
& 0.5787  & 0.6852  \\
\bottomrule
\end{tabular}
\end{center}
\end{table}

\begin{figure}[t]
    \centering
    \includegraphics[width=.5\textwidth]{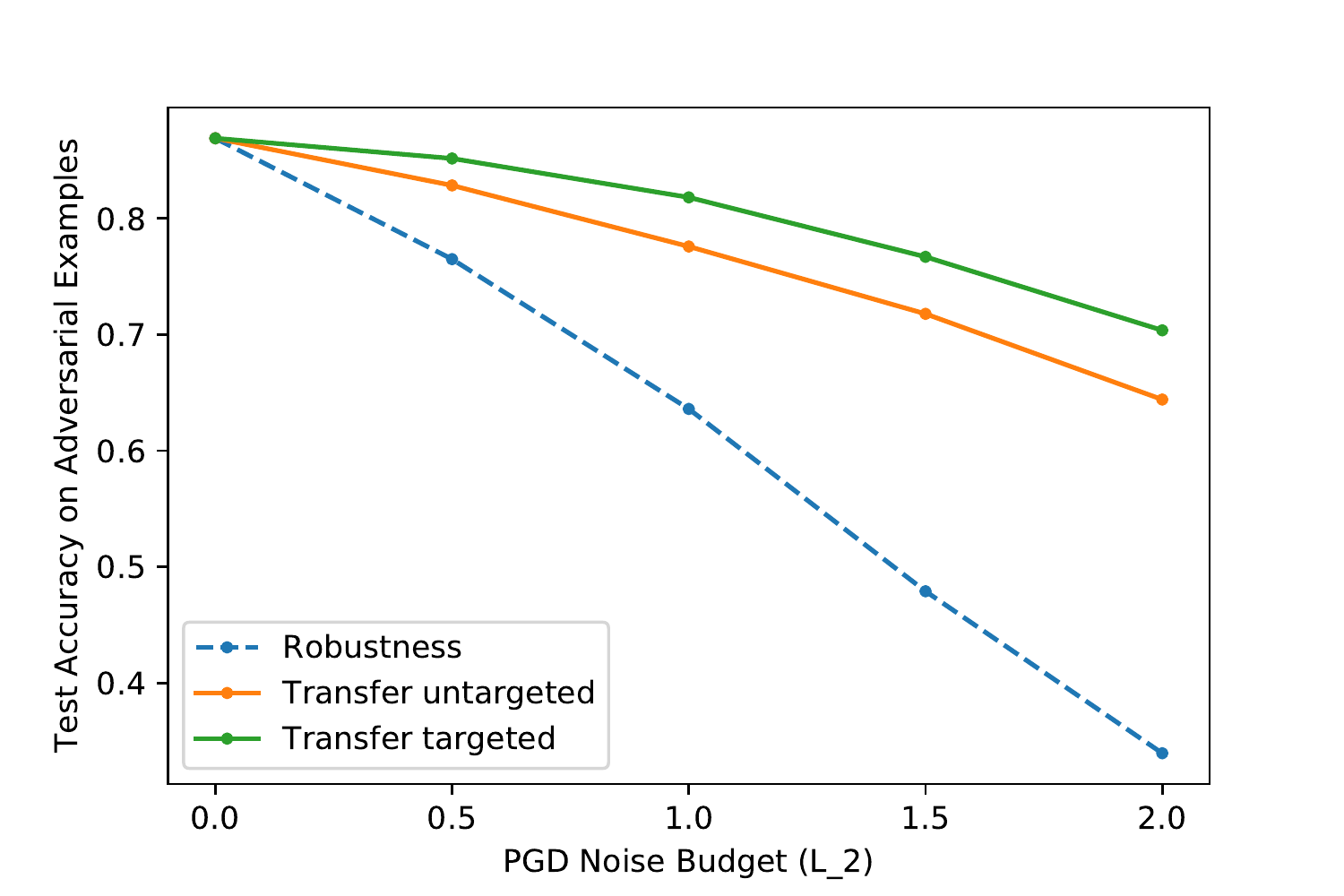}
    \caption{Adverarial examples do not transfer successfully between the multiclass model and \robin. The dotted line corresponds to the robust accuracy of \robin based on the \textsc{Softmax} attack and is indicative of the lack of transferability.}
    \label{fig:transfer}
\end{figure}

We investigate the transferability of adversarial examples from the multiclass benchmark to \robin. One might expect that since the two models are trained over the same dataset to solve the same task, adversarial examples could transfer between the two architectures. Then the multiclass could be effectively used as a surrogate to craft black box attacks, which has been shown be a successful technique.

However, we find that adversarial examples crafted for the multiclass model do not transfer effectively to \robin, indicating that similarity in architecture of the models might be crucial, i.e. the similarity task or common train dataset is not sufficient to guarantee high transferability.

We consider both targeted and untargeted transfer attacks. For the untargeted case, we perform an untargeted attack to the multiclass model and check whether the adversarial example succeeds in fooling \robin. For the targeted case, we first evaluate the predictions of \robin using the clean example. Then, we perform a targeted attack on the multiclass model and choose the target label to be the second highest arm (second largest prediction score) of \robin. Theoretically, this would be the easiest label to attack.

The results of this experiment using ResNet-18 on CIFAR-10 can be found in Table~\ref{tab:transferability}. The model is adversarially trained using PGD10 with $\epsilon = 0.5$ for $\ell_2$ and $\epsilon = 8/255$ for $\ell_\infty$. In Figure~\ref{fig:transfer} we also provide an illustration of the results for $\ell_2$. For reference we also plot the actual robust accuracy of the model that corresponds to the \textsc{Softmax} attack (dotted line). It is evident that the robust accuracy from transfer-based attack is significantly higher than the actual robustness of \robin, indicating that adversarial examples from the multiclass do not transfer very efficiently.

\subsection{Sampling for Class Imbalance} \label{app:setup}

To handle class imbalance while training binary classifiers in \robin, we can downsample negative labels and upsample positive labels so that each training batch has an equal number of positive and negative labels.

We find that overall, sampling improves the clean test accuracy of \robin while still improving robustness. Table~\ref{tab:main-res} in the main paper shows results from \robin using sampling on CIFAR-10. On MNIST, \robin is trained without sampling. In Tables~\ref{tab:resnet-cifar10-app} and \ref{tab:resnet-cifar10-app4} we show results for \robin without and with sampling respectively, for a number of different noise budgets during test time. Note that the clean accuracy ($\epsilon=0$) of \robin improves with sampling especially for higher train $\epsilon$. However, in both cases, \robin still improves upon the robustness of the benchmark when the test noise budget is at least as large as the train budget.

\begin{table*}
  \caption{Comparison of robust accuracy between a standard multiclass classifier (MC) and \robin on CIFAR-10 using ResNet-18, where \robin is trained {\bf without sampling}. Performance across varying test time $\epsilon$ budgets are included as well as the difference between \robin and multiclass (Gap). Consistently \robin outperforms the multiclass when the $\epsilon$ at test time equals the one used at train time. For higher $\epsilon$, \robin improves upon multiclass.}
\label{tab:resnet-cifar10-app}
 \centering
  \begin{tabular}{cc|ccc|ccc}
  \toprule

& & \multicolumn{6}{c}{ResNet-18 CIFAR-10 without sampling}\\
 \toprule
 
& & \multicolumn{6}{c}{Train $\epsilon$}\\

& $\ell_2$ radius  &  \multicolumn{3}{c}{{$\epsilon=0.5$}}   & \multicolumn{3}{c}{{$\epsilon=1.0$}} \\ \hline
\multirow{5}{*}{\rotatebox[origin=c]{90}{Test $\epsilon$}} & &  MC & \robin & Gap & MC & \robin & Gap \\

& $\epsilon = 0.0$ & \textbf{0.9153} & 0.8972 & -0.0181   &     \textbf{0.8765} & 0.8359 & -0.0406     \\
& $\epsilon = 0.5$  & 0.7498 & \textbf{0.7582} & +0.0084   &     \textbf{0.7714} & 0.7474 & -0.024              \\
& $\epsilon = 1.0$  & 0.5057 & \textbf{0.5412} & +0.0355   &     0.629 & \textbf{0.6323} & +0.0033   \\
& $\epsilon = 2.0$ & 0.1236 & \textbf{0.1747} & +0.0511   &     0.3151 & \textbf{0.3746} & +0.0595 \\

& $\epsilon = 3.0$  & 0.0178 & \textbf{0.0308} & +0.013   &     0.1176 & \textbf{0.1656} & +0.048 \\

\midrule
& $\ell_\infty$ radius &
\multicolumn{3}{c}{{$\epsilon=4/255$}}   & \multicolumn{3}{c}{{$\epsilon=8/255$}} \\ 
\hline
\multirow{5}{*}{\rotatebox[origin=c]{90}{Test $\epsilon$}} & &  MC & \robin & Gap & MC & \robin & Gap\\

& $\epsilon = 0.0$ & \textbf{0.8974} & 0.8656 & -0.0318   &     \textbf{0.7659} & 0.7597 & -0.0062     \\
& $\epsilon = 4/255$  & 0.6209 & \textbf{0.6386} & +0.0177   &     0.609 & \textbf{0.6202} & +0.0112                \\
& $\epsilon = 6/255$  & 0.4328 & \textbf{0.4937} & +0.0609   &     0.5408 & \textbf{0.5448} & +0.004    \\
& $\epsilon = 8/255$  & 0.2784 & \textbf{0.353} & +0.0747   &     0.4371 & \textbf{0.4651} & +0.028\\

& $\epsilon = 10/255$  & 0.1629 & \textbf{0.2324} & +0.0695   &     0.3362 & \textbf{0.381} & +0.0448\\

\bottomrule
\end{tabular}
 \end{table*}

\begin{table*}
  \caption{Comparison of robust accuracy between a standard multiclass classifier (MC) and \robin on CIFAR-10 using ResNet-18, where \robin is trained {\bf with sampling}. Performance across varying test time $\epsilon$ budgets are included as well as the difference between \robin and multiclass (Gap). Consistently \robin outperforms the multiclass when the $\epsilon$ at test time equals the one used at train time. For higher $\epsilon$, \robin improves upon multiclass.}
\label{tab:resnet-cifar10-app4}
 \centering
  \begin{tabular}{cc|ccc|ccc}
  \toprule

& & \multicolumn{6}{c}{ResNet-18 CIFAR-10 with sampling}\\
 \toprule
 
& & \multicolumn{6}{c}{Train $\epsilon$}\\

& $\ell_2$ radius  &  \multicolumn{3}{c}{{$\epsilon=0.5$}}   & \multicolumn{3}{c}{{$\epsilon=1.0$}} \\ \hline
\multirow{5}{*}{\rotatebox[origin=c]{90}{Test $\epsilon$}} & &  MC & \robin & Gap & MC & \robin & Gap \\

& $\epsilon = 0.0$ & \textbf{0.9153} & 0.9049 & -0.0104   &     \textbf{0.8902} & 0.8691 & -0.0211    \\

& $\epsilon = 0.5$  & 0.7498 & \textbf{0.7719} & +0.0221  &     \textbf{0.7792} & 0.765 & -0.0142          \\
& $\epsilon = 1.0$  & 0.5057 & \textbf{0.5648} & +0.0591   &     0.6168 & \textbf{0.6359} & +0.0191   \\
& $\epsilon = 2.0$ & 0.1236 & \textbf{0.2109} & +0.0873   &     0.2805 & \textbf{0.3395} & +0.059 \\

& $\epsilon = 3.0$  & 0.0178 & \textbf{0.0399} & +0.0222  &     0.0811 & \textbf{0.1446} & +0.0635 \\

\midrule
& $\ell_\infty$ radius &
\multicolumn{3}{c}{{$\epsilon=4/255$}}   & \multicolumn{3}{c}{{$\epsilon=8/255$}} \\ 
\hline
\multirow{5}{*}{\rotatebox[origin=c]{90}{Test $\epsilon$}} & &  MC & \robin & Gap & MC & \robin & Gap\\
& $\epsilon = 0.0$ & \textbf{0.8974} & 0.8736 & -0.0238   &     0.7659 & \textbf{0.8053}  & +0.0394   \\
& $\epsilon = 4/255$  & 0.6209 & \textbf{0.6259} & +0.005 &     0.6092 & \textbf{0.6395} & +0.0303            \\
& $\epsilon = 6/255$  & 0.4328 & \textbf{0.4673} & +0.0345   &     0.5404 & \textbf{0.5454} & +0.005   \\
& $\epsilon = 8/255$  & 0.2784 & \textbf{0.3334} & +0.055  &     0.4371 & \textbf{0.4484} & +0.0113\\

& $\epsilon = 10/255$  & 0.1629 & \textbf{0.2225} & +0.0596  &     0.336 & \textbf{0.3588} & +0.0228\\

\bottomrule
\end{tabular}
 \end{table*}

\section{Hierarchical Classifiers}
\label{app:stacked}

\begin{figure*}[t]
\begin{center}
\begin{tabular} {cccc}
  \includegraphics[width=.45\textwidth]{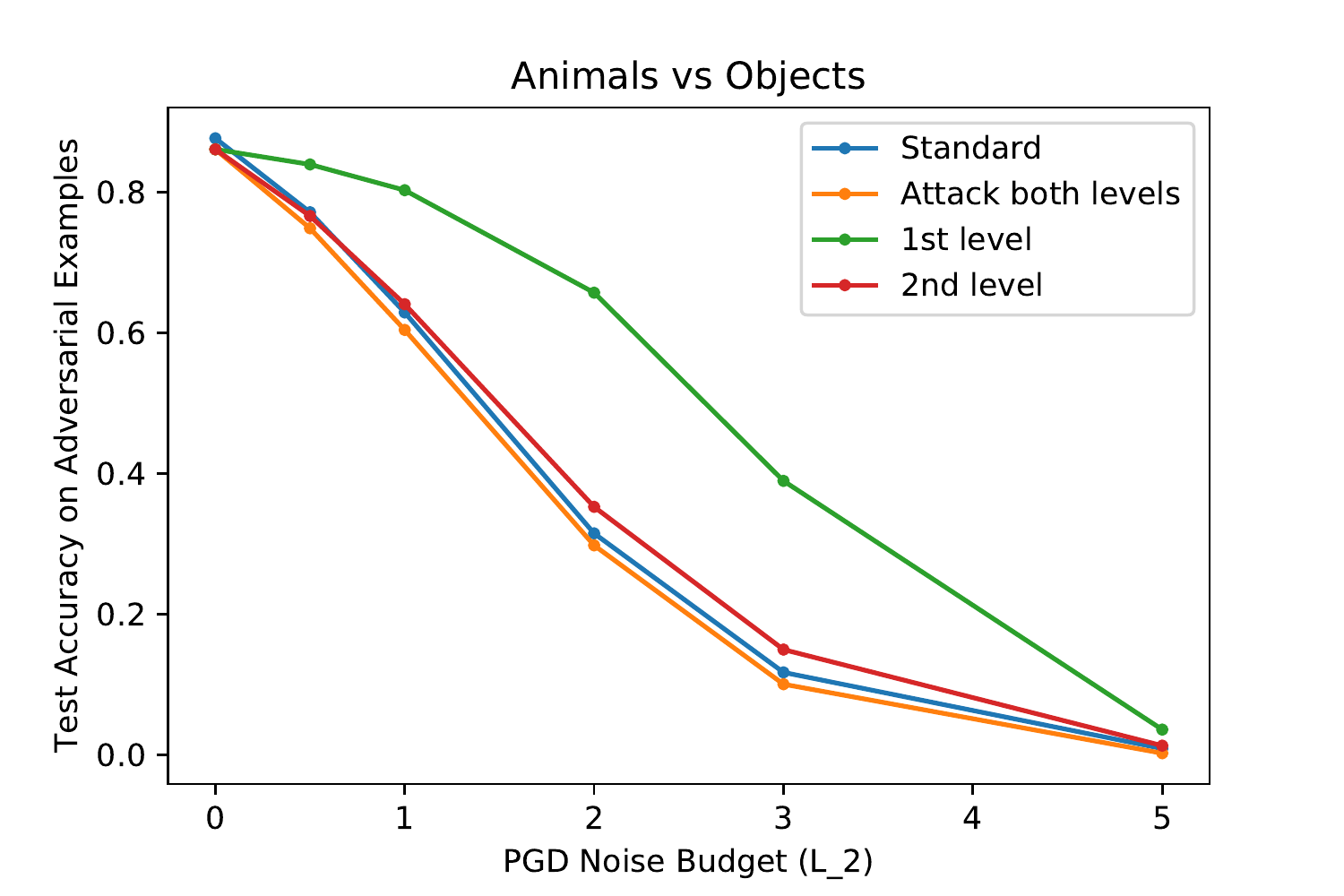} & 
  \includegraphics[width=.45\textwidth]{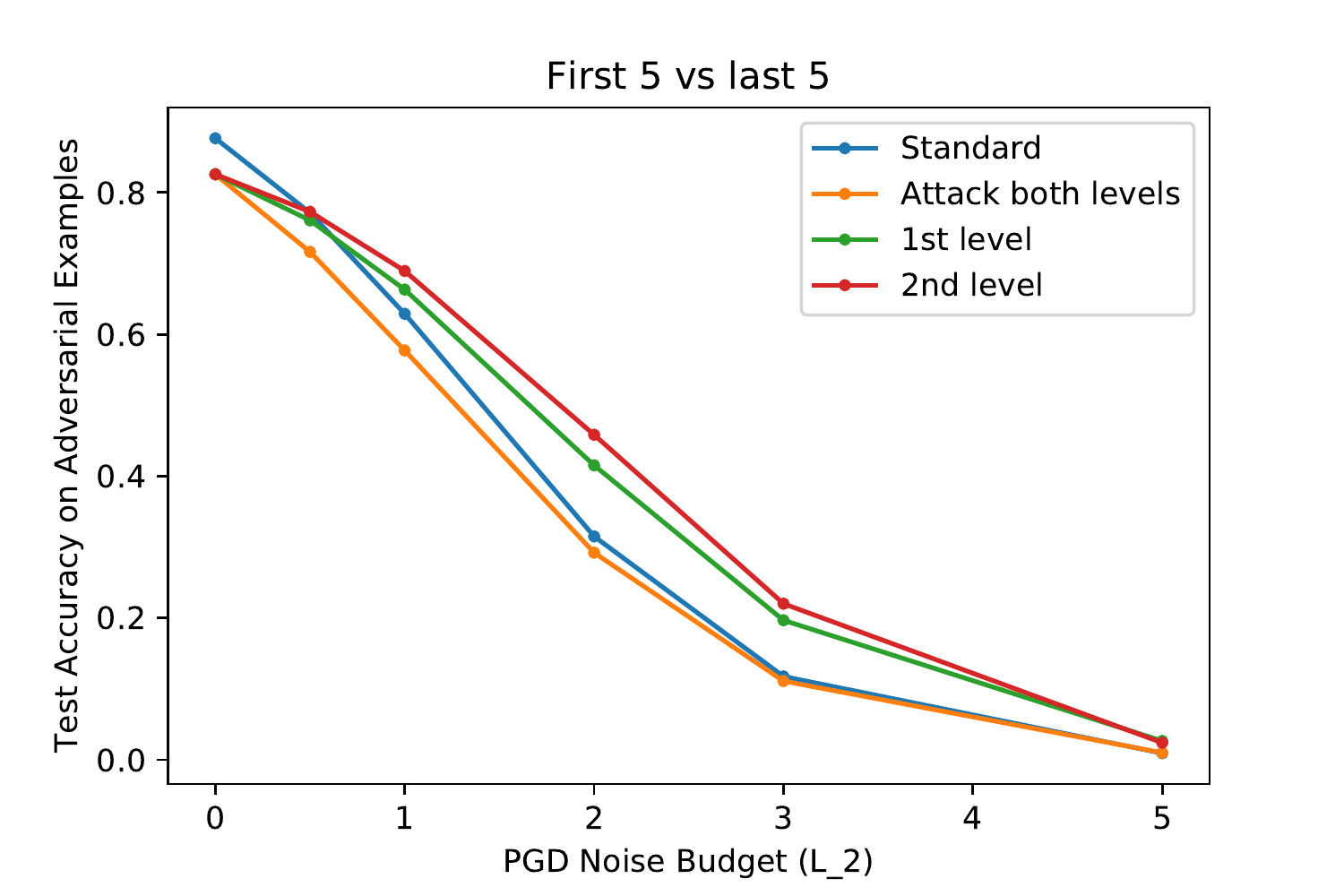} \\
  \end{tabular}
\end{center}
  \caption{\textbf{Left:} ``Animals vs. objects" partition in the first stage. The model is much more robust against attacks that target first stage. \textbf{Right:} ``First 5 vs. last 5" partition in the first stage. This partition is a difficult classification task and hence the model is not robust to attacks on the first stage. Additionally, the total robustness is lower than the case where we have an easier partition in the first stage.}
  \label{fig:stacked}
\end{figure*}

A drawback of \robin (and of other multiclass reduction methods such as one-vs-one) is that it requires training many binary classifiers. One cannot help but wonder whether all these classifiers are truly necessary for robustness. In an alternative (but similar to \robin) approach we attempt to decompose the multiclass classification task into simpler, but more complex than binary, classification problems. 

We train three multiclass models on CIFAR-10 that we aggregate into one hierarchical classifier. The first model performs a ``first-stage" classification where we train a model to predict coarse labels for the points in the dataset, for example animals vs. objects in CIFAR-10. In the ``second-stage", we predict fine labels using a second model to predict 6 categories of animals or a third model which differentiates between 4 categories of objects. We aggregate the model, where we first input the image into the first classifier to select the coarse label with the highest score before deciding to feed the image into either the second or third classifier. Generally, there are as many fine models as the number of coarse labels and each of them is trained only on the subset of the training set that contains points labeled with the respective coarse label.

The question is whether such a model can give similar gains in robustness as \robin. Because of the hierarchical structure, there are multiple approaches to attack such a model:\begin{enumerate*}
    \item attack only the first stage model,
    \item attack only the relevant second stage model,
    \item attack the first stage model and if unsuccessful attack the second stage model.
\end{enumerate*}
\vspace{-.3cm}Again, each of these approaches needs to be coupled with a base attack such as PGD.

We evaluate the robustness of a hierarchical model in comparison to that of a standard multiclass model. We use ResNet-18 for the first and second stage as well as for the multiclass benchmark. All of the models are adversarially trained using PGD10 with $\ell_2$ budget $\epsilon = 0.5$. We try two different first stage splits, i.e. ``animals vs. objects" and ``first 5 vs. last 5" and we present the results in Figure~\ref{fig:stacked}.

We find that even these simple attacks, which are not designed to holistically attack the hierarchical model, but rather target each component separately, are strong enough attacks to push robustness below the robustness of the corresponding multiclass benchmark.

Moreover, we find that the initial partition of the fine labels into coarse ones is important and yields very different robustness results. Specifically, if classification in the first stage is an easy task, e.g. as in the animals vs. objects classification task in CIFAR-10, then we can achieve better robustness results at this level. On the other hand if the first stage classification task is difficult as in the case of first 5 vs. last 5 classes of CIFAR-10, then attacking the first level could be more effective than attacking the models that perform the final classification of finer labels.

Hence, there is no clear separation in the robustness between the first and the second stage. The susceptibility to attacks depends on the partition of fine labels that was selected.

\end{document}